\documentclass[letterpaper]{article} 
\usepackage[preprint]{aaai2027}  
\usepackage[hyphens]{url}  
\usepackage{graphicx} 
\usepackage{natbib}  
\usepackage{caption} 
\usepackage{algorithm}
\usepackage{multirow}
\usepackage[table]{xcolor}
 \usepackage{subcaption} 
\usepackage{latexsym}
\usepackage[T1]{fontenc}
\usepackage[utf8]{inputenc}
\usepackage{microtype}
\usepackage{amsmath,amssymb,amsthm,mathtools}
\usepackage{booktabs}
\usepackage{array}
\usepackage{enumitem}
\usepackage{graphicx}
\usepackage{dblfloatfix}
\usepackage{algorithm}
\usepackage[noend]{algpseudocode}
\usepackage{tabularx}
\usepackage{xcolor}

\algrenewcommand\algorithmicrequire{\textbf{Require:}}
\algrenewcommand\algorithmicensure{\textbf{Return:}}

\newcommand{\method}{\textsc{GUARD}}
\newcommand{\vf}{\textsc{Verified-Faithful}}
\newcommand{\rl}{\textsc{Reject-Leaky}}
\newcommand{\amg}{\textsc{Abstain-Missing-Guards}}
\newcommand{\anc}{\textsc{Abstain-Not-Compatible}}

\newcommand{\Guard}{\mathcal{G}}

\usepackage{newfloat}
\usepackage{listings}
\DeclareCaptionStyle{ruled}{labelfont=normalfont,labelsep=colon,strut=off} 
\floatstyle{ruled}
\newfloat{listing}{tb}{lst}{}
\floatname{listing}{Listing}

\usepackage{booktabs}

\title{Autoformalizing Argumentative Material Inferences}
\author {
    Xin Quan\textsuperscript{\rm 1,\rm 3},
    Reto Gubelmann\textsuperscript{\rm 2},
    Andr\'e Freitas\textsuperscript{\rm 1,\rm 3}\corresponding
}
\affiliations {
    \textsuperscript{\rm 1}Idiap Research Institute, Switzerland\\
    \textsuperscript{\rm 2}University of Zurich, Switzerland\\
    \textsuperscript{\rm 3}Department of Computer Science, University of Manchester, UK\\
    xin.quan@idiap.ch, reto.gubelmann@uzh.ch, andre.freitas@idiap.ch
}

\input{appendix_setup}

\begin{document}

\maketitle

\begin{abstract}
Natural language arguments are compelling before they are formally explicit. A premise supports a claim through defeasible warrants, background commitments, and exception conditions that the text leaves implicit. However, formal verification requires the opposite. Making such arguments machine-checkable requires constructing the missing commitments, not only translating given sentences into logic. Construction, however, carries a risk that translation does not: a system free to add premises can make any claim provable, and a formally valid proof may assert the claim outright, prove it without the original premise, or establish more than the claim itself. We address this problem by formulating autoformalization for argumentative material inference as guard completion, in which non-monotonic material support is turned into monotonic formal inference relative to an explicitly constructed guard set. A completion is accepted only when its proof both passes the theorem prover and survives contrastive tests of premise dependence and claim selectivity. We implement this formulation in \method{}, a neuro-symbolic framework in which LLMs construct and formalize candidate guards, Isabelle/HOL verifies the resulting theories and returns step-level feedback for iterative refinement, and the system abstains when no faithful completion can be reached. Our empirical results on Debatepedia and ARCT using different LLMs demonstrate that \method{} yields significant improvements in verified-faithful inference ($+$35.3, $+$32.9 points) and substantial reductions in leakage ($-$25.9, $-$21.9 points) over the state-of-the-art LLM-driven theorem proving approach. Moreover, we show that the symbolic soft critique and the explicit assumption layer account for most of these gains, with the soft critique also improving the initial validity of the elicited context and reducing the number of iterations required for successful verification.
\end{abstract}

\section{Introduction}

Arguments in natural language are rarely delivered as formal derivations. In debate, explanation, and everyday reasoning, a premise counts as a reason for a claim because readers can recover a warrant, rely on shared background commitments, and leave exception conditions unstated. Argumentation theory has long treated such warrants as part of the layout of reasons \citep{toulminUsesArgument1958,lawrenceArgumentMiningSurvey2020}, while inferentialist accounts describe this support as material inference: inference licensed by the content of the concepts involved and the commitments of the discourse, rather than by logical form alone \citep{brandomMakingItExplicit1994,brandomSayingDoingAnalytic2010,brandomArticulatingReasons2021,hlobil2024reasons}. Material inferences can be deductively valid if the truth of the premise necessitates the truth of the conclusion (as in ``LA is south of San Francisco, therefore San Francisco is north of LA''), or they can be inductively valid if the premise only makes the conclusion plausible, rational, or likely (as in ``the streets are wet, therefore, it has rained''). It is widely accepted that, typically, material inferences can be explicated into formally valid inferences (as in ``LA is south of San Francisco, and whenever A is south of B, B is north of A, therefore San Francisco is north of LA''). However, due to the context-dependence of aspects of meaning of terms and of the commitments of the discourse, philosophers have seen this task of faithfully explicating implicit material inference into explicit formal inference as a formidable obstacle for AI to overcome \citep[Ch. 3]{brandomSayingDoingAnalytic2010}. 

In AI and logic, the structure referred to by inductive material inference is captured by defeasible and non-monotonic reasoning: a conclusion may be warranted in the current information state and later withdrawn when an exception, priority, or counterargument is introduced \citep{reiter1980logic,mccarthy1980circumscription,pollock1987defeasible,dung1995acceptability}. This implicit and defeasible character runs through NLP inference settings. Natural language inference, commonsense reasoning, and argument mining package materially licensed support under labels such as ``entailment'' or ``argument support'' \citep{bowmanLargeAnnotatedCorpus2015,williamsBroadCoverageChallengeCorpus2018a,nieAdversarialNLINew2020,storks2019commonsense,gubelmann2024capturing}, and argument reasoning benchmarks show that the decisive step is often the reconstruction of an implicit warrant rather than the application of a stated rule \citep{habernal-etal-2018-argument,gupta-etal-2024-harnessing}.
 
Making this support formally explicit is the precondition for verification. A justification can be audited only when the commitments it relies on are stated, and plausible-sounding explanations are known to often fail validity or faithfulness \citep{kumar-talukdar-2020-nile,valentino-etal-2021-natural}. A theorem prover enforces exactly this discipline, since consequence is checked against an explicit theory and every assumption, definition, and exception condition a proof relies on must be written down \citep{nipkow2002isabelle}. Recent autoformalization work makes the construction of such theories feasible, with LLMs translating informal statements into proof-oriented representations and revising them under formal feedback \citep{wu2022autoformalization,olausson-etal-2023-linc,han2023automatic,zhang-etal-2024-consistent}, and a closer line of work couples LLMs with theorem provers to verify and iteratively refine natural-language explanations for NLI \citep{quan-etal-2024-enhancing,quan-etal-2024-verification,quan-etal-2025-faithful,quan-etal-2025-peirce}.
 
However, autoformalizing argumentative material inference raises challenges that do not arise when the supporting content is already given. First, unlike explanation verification, where the sentences to formalize are provided in advance \citep{quan-etal-2024-verification}, the commitments that make the premise sufficient must first be constructed, together with the exception conditions under which they fail. Second, once the system is free to add premises, formal validity alone becomes uninformative. A proof can become valid because an added guard states the answer, weakens the target, or makes the original premise irrelevant. Third, solver feedback cannot distinguish a faithful explicitation of the original support from an easier substitute proof, so verification must be paired with explicit checks on premise dependence and claim selectivity.
 
In this paper, we frame autoformalization for material inference as a guard-completion problem. We propose a neuro-symbolic framework \method{} that integrates LLMs with the Isabelle/HOL theorem prover \citep{nipkow2002isabelle} to investigate the following research questions: \textbf{RQ1:} \textit{Can autoformalization close the explicitness gap for non-monotonic argumentative support without distorting the original material relation?} \textbf{RQ2:} \textit{Can autoformalization-based feedback refine hidden warrants into explicit guard assumptions?} \textbf{RQ3:} \textit{How robust and faithful is guard completion across backbone LLMs?}
 
To answer these questions, \method{} first constructs a strengthened context organized around the sources of implicitness that classical argument reconstruction identifies: enthymematic assumptions, definitions and term normalizations, and the exception conditions under which a defeasible warrant is blocked \citep{naess1953interpretation,pollock1987defeasible,gordon2007carneades}. The elicited defeasible rules are integrated into guarded form with explicit abnormality conditions in the style of circumscription \citep{mccarthy1986applications}, which is how non-monotonic support is turned into monotonic inference relative to explicit guards. An LLM-based soft verifier then screens each candidate against four proof obligations. In turn, Isabelle/HOL is adopted as the hard verifier, and its failed proof steps are extracted as feedback for targeted refinement, extending the soft and hard critique design used in explanation refinement \citep{quan-etal-2025-faithful}. Finally, a contrastive faithfulness stage replaces the premise with alternatives that should not support the claim, and the claim with alternatives that the premise does not warrant, and accepts a valid inference only if every perturbed theory becomes unprovable.
 
Our empirical evaluation on the Debatepedia argument corpus \citep{cabrio2013natural} and the Argument Reasoning Comprehension Task (ARCT) \citep{habernal-etal-2018-argument}, conducted with GPT-5.1, Qwen3-Max, DeepSeek-V3.2, and Mistral Medium 3.5 as backbone LLMs, shows that materially supported arguments consistently lack formal validity in their original form, and that \method{} improves the average verified-faithful rate over the strongest baseline by 35.3 points on Debatepedia and 32.9 points on ARCT, while reducing leakage by 25.9 and 21.9 points respectively. Removing the soft symbolic critique or the assumption layer lowers the verified-faithful rate by up to 40.5 and 31.5 points, and an extended-budget probe shows that most abstentions are missing guards rather than incompatible arguments.

To summarize, this paper makes three contributions. First, we introduce \method{}, an LLM+Isabelle/HOL framework that formulates material-inference autoformalization as guard completion, constructing explicit background context, selecting proof-relevant guards, and checking contrastive faithfulness. Second, we evaluate \method{} on Debatepedia and ARCT across four backbone LLMs, showing average gains of 35.3 and 32.9 points in verified-faithful rate over the strongest baseline and reductions of 25.9 and 21.9 points in leakage. Third, our diagnostic analysis of the explicitation process attributes drops of up to 40.5 and 31.5 verified-faithful points to the soft symbolic critique and the assumption layer, confines residual leakage to premise-side failures, and traces 63.5 to 74.3 percent of the abstained instances to missing guards rather than incompatible arguments.

\begin{figure*}[t]
\centering
\includegraphics[width=0.98\textwidth]{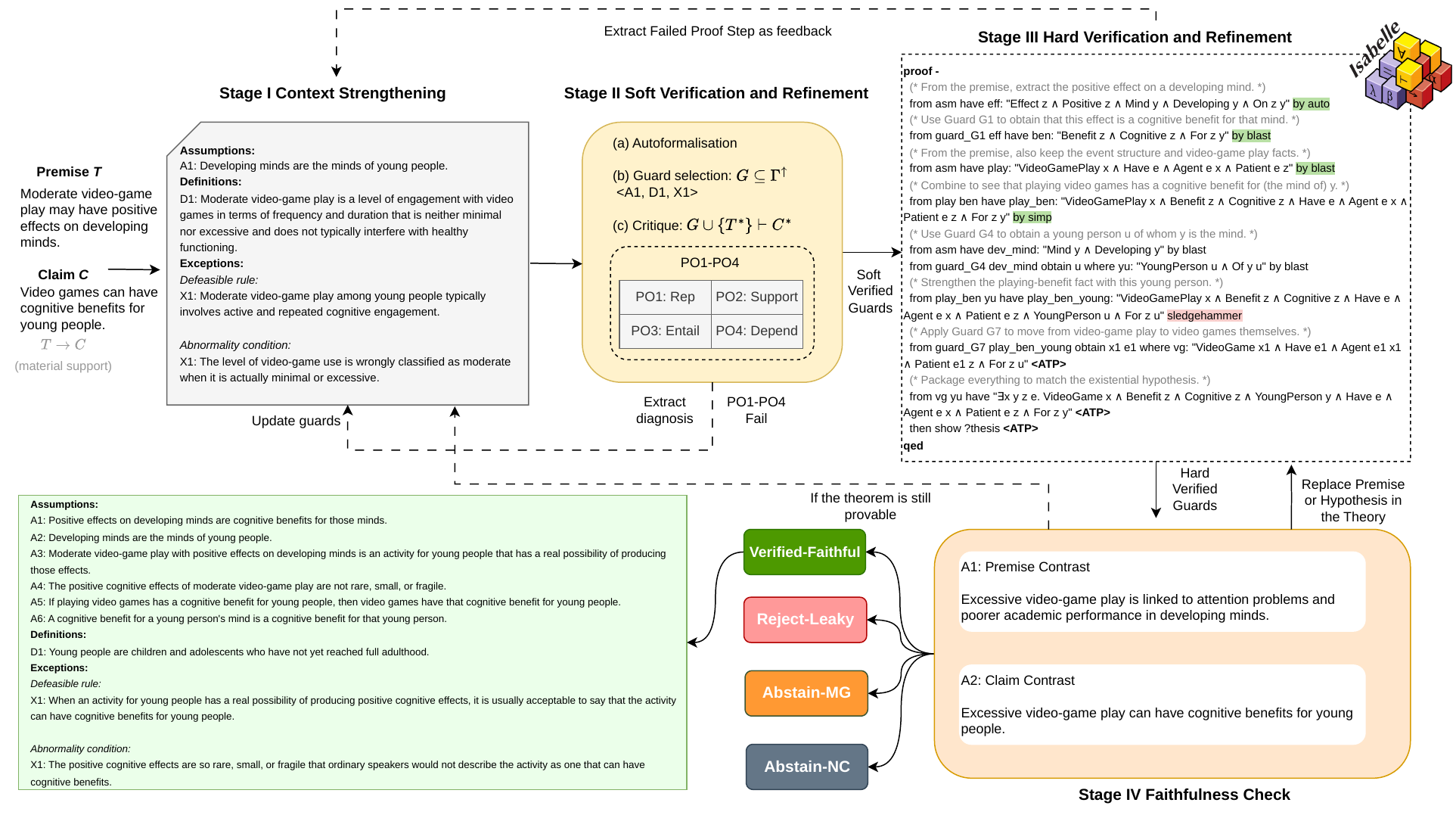}
\caption{Overview of \method{}. Starting from a premise-claim pair $(T,C)$, the framework constructs a strengthened context, autoformalizes the selected candidate guards and iteratively combines soft verification, hard theorem proving and contrastive faithfulness checks until one of the four final statuses is reached.}
\label{fig:framework}
\end{figure*}

\section{Problem Definition}
\label{sec:problem_definition}

In this paper, we define an argument as a premise-claim pair $(T, C)$ in which the premise $T$ provides a materially licensed reason for the claim $C$. Our goal is to construct an explicit strengthened context $\Gamma^{\uparrow}$ and to identify a set of guards $\Guard \subseteq \Gamma^{\uparrow}$ that turns this material support into formal support, such that $\Guard \cup \{T\} \models C$ holds.

We leverage an LLM-based soft verifier and an external theorem prover to systematically verify these entailments through an automated process. Specifically, given the set of input sentences $S = \{T, C\} \cup \Guard$, we aim to construct a set of logical forms $\phi = \{\Phi(s) \mid s \in S\}$, where $\Phi$ is the autoformalization process that converts natural language sentences into symbolic representations. We denote the logical forms of the premise and claim by $T^{\ast} = \Phi(T)$ and $C^{\ast} = \Phi(C)$, respectively, and write $g^{\ast} = \Phi(g)$ for each guard $g \in \Guard$. From these logical forms, we construct a theory $\Theta = (A, \tau)$, where $A = \{g^{\ast} \mid g \in \Guard\}$ is the set of axioms derived from formalizing the guards, and $\tau$ is the theorem to be proven, with $T^{\ast}$ as its assumption and $C^{\ast}$ as its proof goal. Proving $\tau$ thus establishes $A \cup \{T^{\ast}\} \vdash C^{\ast}$. If the theorem prover derives a valid proof of $\tau$ under $A$, we conclude that $\Guard$ formally licenses the inference from $T$ to $C$ under $\Gamma^{\uparrow}$.

Otherwise, a refinement controller uses the failed proof steps as feedback
and iteratively refines the guard set, updating the state
\[
s_k = \langle \Gamma^{\uparrow}_k,\, \Guard_k,\, T^{\ast},\, C^{\ast} \rangle,
\]
generating a refined guard set $\Guard_{k+1}$, until a valid proof is found or a predefined refinement budget is exhausted.

\section{Methodology}
 
\subsection{Stage I: Context Strengthening}
Stage~I is grounded in classical argument reconstruction: everyday arguments are materially valid in virtue of implicit background commitments and become formally assessable once these commitments are made explicit \citep{sellars1953inference,brandomMakingItExplicit1994}. Starting from the input pair $(T,C)$, we construct an explicit strengthened context $\Gamma^\uparrow = A \cup \Delta \cup X$, intended as a task-specific explicit witness to the otherwise implicit discourse context, with each component elicited by a dedicated harvester under shared contextual parameters (e.g., domain, jurisdiction, and timebound). $A$ contains \emph{assumptions} in the sense of enthymeme completion \citep{vaneemeren2004systematic}, including normalcy conditions, domain restrictions, and ceteris-paribus clauses. Each assumption is typed and linked to the premise-claim edge it supports. $\Delta$ contains \emph{definitions and term-normalization facts} in the spirit of making it more precise \citep{naess1953interpretation,walton2001persuasive}, retrieved from authoritative sources and normalized into canonical single-sentence forms for formalization. $X$ contains \emph{exception management}, following defeasible reasoning \citep{toulminUsesArgument1958,pollock1987defeasible,gordon2007carneades}. Based on the elicited assumptions and definitions, we organize the latter as local defeasible rules over intermediate concepts and pair each rule with the exception conditions under which it is blocked. Stage~I therefore builds the strengthened background from which a step-specific guard set $\Guard \subseteq \Gamma^\uparrow$ will later be selected. The details for each harvester are described in the Appendix \ref{sec:prompts}.
 
\subsection{Stage II: Soft Verification and Refinement}

\paragraph{Autoformalization.}
Stage~II begins by applying the autoformalization function $\Phi$ to the context $\Gamma^\uparrow$ elicited in Stage~I. Following \citet{quan-etal-2024-verification}, the function maps each natural language sentence $s$ to a logical form $\Phi(s)$ using Neo-Davidsonian event semantics \citep{parsons1990events} coupled with first-order logic. The resulting logical forms are then aligned with the Isabelle/HOL style. For example, the sentence \emph{``Child beauty pageants can teach children to base self-worth on physical appearance''} can be formalized as follows:
\[
\begin{aligned}
\forall x\,y\,e_1.\;
  & \mathit{ChildBeautyPageant}(x) \wedge \mathit{Child}(y) \wedge \mathit{Teach}(e_1)\\
  & \wedge \mathit{Agent}(e_1,x) \wedge \mathit{Patient}(e_1,y) \longrightarrow\\
\exists z\,w\,e_2.\;
  & \mathit{SelfWorth}(z) \wedge \mathit{PhysicalAppearance}(w) \wedge\\ \mathit{Base}(e_2)
  & \wedge \mathit{Agent}(e_2,y) \wedge \mathit{Patient}(e_2,z) \wedge \mathit{On}(z,w).
\end{aligned}
\]
Here, the verb \emph{teach} is represented as the event $e_1$, with the pageant as its agent and the child as its patient, while the taught behavior of basing self-worth on physical appearance is captured by a second event $e_2$. This predicate-argument structure keeps the logical form close to the surface form of the sentence and preserves its information content during translation. Each assumption and definition in $\Gamma^\uparrow$ is then converted into an axiom. The exception structures elicited in Stage~I are then integrated into guarded defeasible rules, where each rule family $d$ is guarded by an abnormality predicate in the style of circumscription \citep{mccarthy1986applications}: $\forall k.\,\bigl(\varphi_d(k) \wedge \neg\mathit{Ab}_d(k) \rightarrow \psi_d(k)\bigr)$, $\mathit{Ab}_d(k) \leftrightarrow \chi_{d,1}(k) \vee \cdots \vee \chi_{d,n}(k)$, in which $\varphi_d$ and $\psi_d$ denote the body and head of the rule and each $\chi_{d,i}$ describes an exception condition. The biconditional is replaced with a one-directional implication if the exception list is open-ended. The details are described in the Appendix \ref{sec:autoformalisation} and Appendix \ref{sec:prompts}.

\paragraph{Guard selection.}
We define an LLM-based critic function $f_{\mathrm{guard}}$ to identify the contextual support needed for the inference from $T^\ast$ to $C^\ast$. The function uses the formalised logical forms of $\Gamma^\uparrow$ to select a step-specific guard set $\Guard = f_{\mathrm{guard}}(\Gamma^\uparrow, T^\ast, C^\ast) \subseteq \Gamma^\uparrow$, with formal representation $G = \{\, g^\ast \mid g \in \Guard \,\}$. Selection aims to make the guards jointly sufficient to infer $C^\ast$ from $T^\ast$ and the resulting candidate is then assessed against PO1 to PO4.
 
\paragraph{Soft verification: PO1-PO4.}
The framework then performs a soft critique of $\langle G, T^\ast, C^\ast \rangle$ over four proof obligations:
\begin{description}[nosep,leftmargin=1em]
\item[PO1 (Representation correctness)] The formal representation is syntactically well-formed, the types, arities, variable bindings, and semantic roles are coherent, and predicate normalization is consistent across the premise, the claim, and the guards.
\item[PO2 (Support adequacy)] Each guard in $G$ is traceable to an item in $\Gamma^\uparrow$. The guards jointly cover the missing inferential material between $T^\ast$ and $C^\ast$ and make the relevant exception conditions explicit where needed.
\item[PO3 (Strict support)] The guards make the premise formally support the claim: $G \cup \{T^\ast\} \vdash C^\ast$.
\item[PO4 (Dependence)] The proof genuinely depends on both the premise and every guard: (a)~$G \nvdash C^\ast$; (b)~$(G \setminus \{g\}) \cup \{T^\ast\} \nvdash C^\ast$ for all $g \in G$.
\end{description}

\paragraph{Soft refinement.}
We define an LLM-based soft refinement function $f_{\mathrm{refine}}^{\mathrm{soft}}$ that applies targeted updates to the current candidate based on the soft critique feedback. For PO1 failures, the function corrects the affected formal representations. A failure of PO2 requires expanding or refining $\Gamma^\uparrow$ by adding missing contextual items or correcting existing ones. The affected items are then formalised and $\Guard$ is reselected. When PO3 fails, refinement updates the strengthened context and the formalised guard candidates to establish formal support from $T^\ast$ to $C^\ast$. PO4 failures are addressed by removing padded guards or refining the strengthened context, with the aim of making the proof depend on the premise and on each selected guard. The updated candidate is reassessed against PO1 to PO4, with a maximum of 10 soft refinement iterations.
 
\subsection{Stage III: Hard Verification and Refinement}
 
\paragraph{Hard verification.}
Following \citet{quan-etal-2024-verification}, we construct an Isabelle/HOL theory \citep{nipkow2002isabelle} to formally verify the selected guards. Each formal guard $g^\ast \in G$ is converted into an axiom, while the premise and the claim are converted into the theorem to be proven, where $T^\ast$ constitutes the assumption clause and $C^\ast$ the proof goal. A successful proof therefore establishes $G \cup \{T^\ast\} \vdash C^\ast$. An example of the constructed theory is shown in Figure~\ref{fig:framework}.

\paragraph{Hard refinement.}
If the theorem prover fails to prove a proof step, we extract the failed step together with the applied axioms as external feedback for an LLM-based hard refinement function $f_{\mathrm{refine}}^{\mathrm{hard}}$. Depending on the diagnosis, it corrects the affected formal representations or refines the logical error in the strengthened context $\Gamma^\uparrow$. Guard selection is then updated when needed, and the resulting theory is checked by the theorem prover in the next iteration.
 
\subsection{Stage IV: Faithfulness Check}
A formally verified proof is still not sufficient. Over-broad guards may prove the claim without using the premise, or they may establish more than the claim itself. Stage~IV therefore applies two contrastive tests to each verified candidate.
 
\paragraph{A1: Premise contrast.}
A1 keeps the claim and replaces the premise with alternatives that should not support it. The challenge set $\mathcal{T}'$ contains the empty-premise case $T' = \varnothing$ and a same-topic counter-premis is proposed by our framework. For example, given the premise \emph{``GM crops are disease-resistant and therefore increase yield''} and the claim \emph{``GM food is beneficial,''} a generated counter-premise is \emph{``GM agriculture pushes small traditional farming communities out of business.''} A1 therefore measures \emph{premise sensitivity}: whether the proof genuinely depends on the original premise.
 
\paragraph{A2: Claim contrast.}
A2 keeps the premise and replaces the claim with a nearby alternative that the premise does not warrant, collected in a challenge set $\mathcal{C}'$. The alternative is proposed by the framework as a cross-subtopic claim, i.e., a claim on the same topic whose content the premise does not address. For the same example, the generated alternative is \emph{``GM food is safe for consumers,''} since the premise concerns yield rather than consumer safety. A2 therefore measures \emph{claim selectivity}: whether the proof establishes only the intended claim. We autoformalize each probe and substitute it into the verified theory. The candidate passes the faithfulness check only if none of the substituted theories remains provable: $\forall T' \in \mathcal{T}'.\; G \cup \{T'^\ast\} \nvdash C^\ast$, $\forall C' \in \mathcal{C}'.\; G \cup \{T^\ast\} \nvdash C'^\ast$.
 
\paragraph{Faithfulness refinement.}
We define an LLM-based faithfulness refinement function $f_{\mathrm{refine}}^{\mathrm{faith}}$ to address leakage detected by the contrastive tests. If a substituted theory remains provable, the function takes the leaking theory and a leak report as input and produces a complete updated guard set. A guard whose consequent asserts the claim unconditionally is made conditional on the content supplied by the premise. The function also narrows over-broad guards to exactly what the claim requires. The original entailment must remain provable under the updated guards. We set a maximum of 5 iterations for faithfulness refinement.
 
\paragraph{Outputs.}
The pipeline returns one of four verdicts. A candidate is labelled \vf{} when hard verification succeeds and both contrastive tests pass, and \rl{} when the entailment is formally verified but A1 or A2 still fails after refinement. If no verified candidate is obtained within the refinement budget, we distinguish two outcomes. \amg{} indicates that admissible repairs, such as unresolved assumptions, definitions, exception conditions, or alternative guard choices, are still open when the budget is exhausted. \anc{} indicates that no admissible repair remains, as the remaining options would require altering the claim itself or would yield only leaky or irrelevant proofs.

\begin{table*}[t]
\centering
\scriptsize
\setlength{\tabcolsep}{3.5pt}
\renewcommand{\arraystretch}{1.0}
\begin{tabular}{ll cccccc cccccc}
\toprule
& & \multicolumn{6}{c}{Debatepedia} & \multicolumn{6}{c}{ARCT} \\
\cmidrule(lr){3-8} \cmidrule(lr){9-14}
Backbone & Method & SP $\uparrow$ & VFR $\uparrow$ & Leak $\downarrow$ & A1 $\downarrow$ & A2 $\downarrow$ & \#Iter & SP $\uparrow$ & VFR $\uparrow$ & Leak $\downarrow$ & A1 $\downarrow$ & A2 $\downarrow$ & \#Iter \\
\midrule
\multirow{5}{*}{DeepSeek-V3.2}
& Direct    & 0.00 & 0.00 & -- & -- & -- & --  & 0.00 & 0.00 & -- & -- & -- & -- \\
& CoT       & 14.17 & 11.74 & \underline{17.14} & \underline{11.43} & 5.71 & --  & 16.00 & 12.50 & 21.88 & 15.63 & 9.38 & -- \\
& Toulmin   & 10.93 & 8.10 & 25.93 & 18.52 & 14.81 & --  & 17.00 & 14.00 & \underline{17.65} & \underline{14.71} & \underline{2.94} & -- \\
& PEIRCE    & \underline{28.34} & \underline{19.03} & 32.86 & 30.00 & \underline{4.29} & 5.86 & \underline{31.00} & \underline{23.00} & 25.81 & 24.19 & 6.45 & 4.15 \\
\rowcolor{gray!10}
& \method{} & \textbf{51.82} & \textbf{48.58} & \textbf{6.25} & \textbf{6.25} & \textbf{0.78} & 3.72 & \textbf{55.50} & \textbf{53.50} & \textbf{3.60} & \textbf{3.60} & \textbf{0.00} & 3.82 \\
\midrule
\multirow{5}{*}{Qwen3-Max}
& Direct    & 0.00 & 0.00 & -- & -- & -- & --  & 0.00 & 0.00 & -- & -- & -- & -- \\
& CoT       & 12.96 & 8.10 & 37.50 & \underline{25.00} & 18.75 & --  & 18.00 & 13.00 & 27.78 & 22.22 & 11.11 & -- \\
& Toulmin   & 14.98 & 10.53 & \underline{29.73} & 27.03 & \underline{8.11} & --  & 17.00 & 13.50 & \underline{20.59} & \underline{17.65} & 5.88 & -- \\
& PEIRCE    & \underline{34.82} & \underline{24.29} & 30.23 & 26.74 & 8.14 & 5.94 & \underline{42.00} & \underline{30.00} & 28.57 & 25.00 & \underline{3.57} & 3.27 \\
\rowcolor{gray!10}
& \method{} & \textbf{59.92} & \textbf{56.68} & \textbf{5.41} & \textbf{5.41} & \textbf{0.00} & 3.85 & \textbf{63.00} & \textbf{61.00} & \textbf{3.17} & \textbf{3.17} & \textbf{0.00} & 3.21 \\
\midrule
\multirow{5}{*}{Mistral Medium 3.5}
& Direct    & 0.00 & 0.00 & -- & -- & -- & --  & 0.00 & 0.00 & -- & -- & -- & -- \\
& CoT       & 14.98 & 10.53 & 29.73 & 21.62 & 10.81 & --  & 19.50 & 12.00 & 38.46 & 25.64 & 15.38 & -- \\
& Toulmin   & 18.22 & 13.77 & \underline{24.44} & \underline{17.78} & 11.11 & --  & 16.00 & 12.50 & 21.88 & \underline{18.75} & 6.25 & -- \\
& PEIRCE    & \underline{38.06} & \underline{26.32} & 30.85 & 27.66 & \underline{6.38} & 5.83 & \underline{46.00} & \underline{36.50} & \underline{20.65} & 20.65 & \underline{5.43} & 5.62 \\
\rowcolor{gray!10}
& \method{} & \textbf{64.37} & \textbf{62.75} & \textbf{2.52} & \textbf{2.52} & \textbf{0.00} & 3.74 & \textbf{66.00} & \textbf{64.50} & \textbf{2.27} & \textbf{2.27} & \textbf{0.76} & 3.25 \\
\midrule
\multirow{5}{*}{GPT-5.1}
& Direct    & 0.00 & 0.00 & -- & -- & -- & --  & 0.00 & 0.00 & -- & -- & -- & -- \\
& CoT       & 24.29 & 19.43 & 20.00 & \underline{13.33} & 10.00 & --  & 25.50 & 18.00 & 29.41 & 19.61 & 15.69 & -- \\
& Toulmin   & 24.70 & 20.24 & \underline{18.03} & 18.03 & \textbf{0.00} & --  & 21.00 & 17.50 & \underline{16.67} & \underline{11.90} & \underline{4.76} & -- \\
& PEIRCE    & \underline{45.34} & \underline{33.60} & 25.89 & 17.86 & \underline{9.82} & 5.21 & \underline{51.00} & \underline{39.00} & 23.53 & 19.61 & 5.88 & 4.07 \\
\rowcolor{gray!10}
& \method{} & \textbf{78.14} & \textbf{76.52} & \textbf{2.07} & \textbf{2.07} & \textbf{0.00} & 3.78 & \textbf{82.50} & \textbf{81.00} & \textbf{1.82} & \textbf{1.82} & \textbf{0.00} & 3.13 \\
\bottomrule
\end{tabular}
\caption{Main comparison across four backbone LLMs on Debatepedia and ARCT. `SP' denotes the solver-pass rate. `VFR' denotes verified-faithful rate, the number of instances whose proof additionally passes the contrastive faithfulness checks. `Leak' represents the solver-passed cases rejected by the faithfulness checks. `A1' and `A2' shows the number of solver-passed cases that fail the premise-contrast and the claim-contrast test, respectively. `\#Iter' indicates the average iteration required to refine the proposed context information to be successfully verified by the theorem prover. The best result per backbone and column is in bold and the second best is underlined.}
\label{tab:main-results}
\end{table*}

\section{Experimental Setup}
 
\paragraph{Datasets and Models.}
We select two argumentation datasets for evaluation: a topic-balanced derivative of the extended Debatepedia corpus \citep{cabrio2013natural} and the Argument Reasoning Comprehension Task (ARCT) \citep{habernal-etal-2018-argument}. The extended Debatepedia resource contains argument pairs drawn from 19 online-debate topics. We construct a balanced evaluation set of 247 premise-claim pairs, with exactly 13 pairs per topic. ARCT contains reason-claim pairs in which the reason supports the claim only through an implicit warrant. We randomly sampled 200 instances from ARCT and treat each reason-claim pair as a premise-claim pair. The dataset selection details are provided in
Appendix~\ref{sec:dataset-details}.
 
We conducted experiments with four backbone LLMs within the proposed framework: GPT-5.1 \citep{openaiGPT512025}, Qwen3-Max \citep{qwenTeamQwen3Max2025}, DeepSeek-V3.2 \citep{deepseekV32025}, and Mistral Medium 3.5 \citep{mistralMedium352026}. We employed Isabelle/HOL \citep{nipkow2002isabelle} as the logical solver. Backbone selection and LLM implementation settings are detailed in Appendix~\ref{sec:llm-implementation}.
 
\paragraph{Baselines.}
We compare against four baselines. (1)~\textbf{Direct autoformalisation}: map $(T,C)$ directly into $(T^\ast,C^\ast)$ and use the solver to find a proof without adding explicit context. (2)~\textbf{1-shot CoT}: use chain-of-thought prompting \citep{wei2022chain,wang2023selfconsistency} with one in-context example to generate an ordered chain of two to four free-text background sentences that makes the implicit reasoning from the premise to the hypothesis explicit, and then formalise the generated background contextual information. (3)~\textbf{Toulmin-style 1-shot explication}: follow \citet{gupta-etal-2024-harnessing} and prompt the LLM to explicate the implicit argument in Toulmin terms (i.e., a qualified warrant, its backing, an additional bridging sentence when needed, and a rebuttal condition). The generated Toulmin context is then used as the added bridging text for downstream formalisation and verification. (4)~\textbf{PEIRCE}: use the neuro-symbolic refinement framework of \citet{quan-etal-2025-peirce}, which supplements premise-side explanatory sentences in response to logical verification results so that the premise can prove the hypothesis. In our comparison setting, these explanatory sentences are treated as added context. We set the maximum number of iterations for theorem-prover-based verification and refinement to 10 for both PEIRCE and our approach. The baseline selection rationale and comparison details are described in Appendix~\ref{sec:baseline-details}.
 
\section{Results and Evaluation}

Table~\ref{tab:main-results} reports the main comparison. For CoT and Toulmin, the backbone LLM first generates the bridging context, which is then autoformalised into logical forms and formalised as an Isabelle/HOL theory. Direct, CoT, and Toulmin are verified in a single pass over the generated context, whereas PEIRCE and \method{} refine their candidates iteratively with prover feedback. All methods share the same autoformalization process, whose faithfulness is evaluated in the Appendix \ref{sec:autoformalisation-evaluation}. Every solver-passed case, regardless of method, is submitted to the contrastive faithfulness checks introduced above, from which the Leak, A1, and A2 columns are derived.
 
\textbf{Materially supported arguments lack formal validity and \method{} consistently converts them into formally verified inferences.}
Direct autoformalization fails on every instance: under all four backbones, a solver-pass rate of 0.00 stands on both datasets. The support rests on commitments the text never states, and translation alone cannot recover them. Single-pass explication helps little. CoT and Toulmin generate bridging sentences that spell out the hidden warrant, and the resulting arguments read as complete, yet once formalized they reach only 10.93 to 25.50 SP and end with verified-faithful rates between 8.10 and 20.24. With failed proof steps fed back to refinement function, PEIRCE roughly doubles the single-pass SP under every backbone, to between 28.34 and 51.00: validity accumulates through revision against solver feedback rather than arriving in one shot. \method{} keeps the same iterative loop but directs it at explicitly constructed guards, and the combination performs best throughout. It obtains the highest SP and verified-faithful rate (VFR) for every backbone on both datasets, with an average VFR of 61.13 on Debatepedia and 65.00 on ARCT, 35.3 and 32.9 points above PEIRCE, and it converges within 3.13 to 3.85 hard iterations on average.
\begin{table}[t]
\centering
\scriptsize
\setlength{\tabcolsep}{5pt}
\renewcommand{\arraystretch}{1.0}
\begin{tabular}{l cc cc}
\toprule
& \multicolumn{2}{c}{Debatepedia} & \multicolumn{2}{c}{ARCT} \\
\cmidrule(lr){2-3} \cmidrule(lr){4-5}
Method & Empty $\downarrow$ & Ctr.-only $\downarrow$ & Empty $\downarrow$ & Ctr.-only $\downarrow$ \\
\midrule
CoT       & 15.36 & 2.48 & 19.99 & 0.78 \\
Toulmin   & 18.18 & 2.16 & 14.28 & 1.47 \\
PEIRCE    & 25.03 & 0.53 & 22.36 & \textbf{0.00} \\
\rowcolor{gray!10}
\method{} & \textbf{4.06} & \textbf{0.00} & \textbf{2.72} & \textbf{0.00} \\
\bottomrule
\end{tabular}
\caption{Decomposition of the A1 failures in Table~\ref{tab:main-results} by probe, averaged over the four backbone LLMs. `Empty' denotes candidates whose guards prove the claim with no premise at all, and `Ctr.-only' candidates that fail only once the counter-premise is substituted.}
\label{tab:probe-decomposition}
\end{table}

\textbf{LLM-driven refinement improves provability by integrating additional context but can make the original premise unnecessary. \method{} reduces this leakage through explicit checks on premise dependence.}  Averaged over the four backbones, PEIRCE reaches a solver-pass rate of 36.64\% on Debatepedia and 42.50\% on ARCT, more than doubling the prompting baselines, yet its leakage exceeds Toulmin's on both datasets (29.96\% against 24.53\%, and 24.64\% against 19.20\%). Refining toward provability therefore inflates validity faster than faithfulness: when the solver only rewards a closed proof, the easiest repair is to strengthen the added context until it carries the proof on its own. Table~\ref{tab:probe-decomposition} exposes the mechanism. PEIRCE's leaked candidates fail almost exclusively under the empty premise (25.03\% and 22.36\%, with counter-only rates of 0.53\% and 0.00\%): the failure mode is bypassing the premise, not contradicting it, as the refined sentences never turn against the original premise or fire on an opposing one, they simply make the premise unnecessary. \method{} runs the same LLM-prover loop but blocks this path by construction, through the dependence obligation PO4 during soft verification and the empty-premise probe afterwards, and its leakage falls to 4.06\% and 2.72\%.
 
\begin{figure}[t]
    \centering
    \begin{subfigure}[t]{0.48\linewidth}
        \centering
        \includegraphics[width=\linewidth]{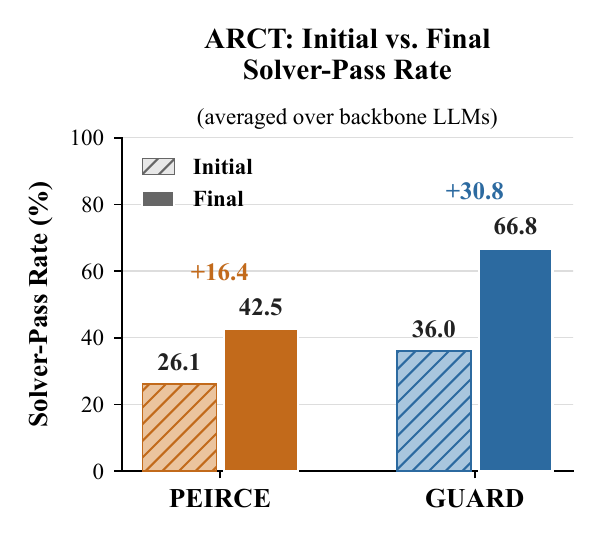}
        \label{fig:gain_compare_arct}
    \end{subfigure}
    \hfill
    \begin{subfigure}[t]{0.48\linewidth}
        \centering
        \includegraphics[width=\linewidth]{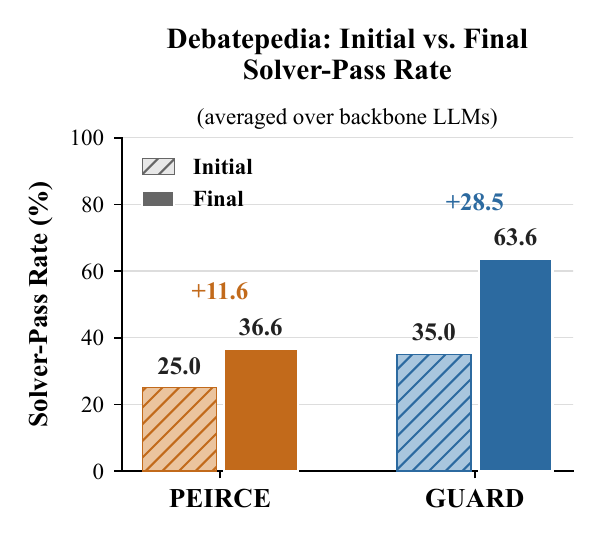}
        \label{fig:gain_compare_debatepedia}
    \end{subfigure}
    \caption{Initial and final solver-pass rates of PEIRCE and \method{} on Debatepedia and ARCT, averaged over the four backbone LLMs.}
    \label{fig:gain}
\end{figure}
 
\textbf{The integration of LLM-based verification and refinement substantially improves the formal validity of the elicited context.} Figure~\ref{fig:gain} compares the initial and final solver-pass rates of PEIRCE and \method{}, where the initial rate counts the instances whose first submitted theory is already proved. Stage II subjects the elicited context to a formal check before the prover ever sees it: the framework critiques the candidate guards at the level of their logical forms against the four proof obligations, and repairs malformed representations, missing support, and padded guards at the same symbolic level. The effect is visible at the first prover call. Although both methods start from the same premise-claim pairs, 35.0\% of \method{}'s first theories are already valid on Debatepedia and 36.0\% on ARCT, against 25.0\% and 26.1\% for PEIRCE, which submits its elicited sentences to the solver directly; on Debatepedia, \method{}'s first submission already comes within 1.6 points of the rate PEIRCE reaches after its full refinement budget. The same symbolic-level refinement also makes the subsequent hard loop more productive: because the theories that reach Isabelle/HOL are structurally sound, a failed proof step localizes a genuine logical gap whose diagnosis maps onto an obligation-specific repair, and \method{} gains 28.6 points on Debatepedia and 30.8 points on ARCT from hard refinement, against 11.6 and 16.4 for PEIRCE, in fewer hard iterations.

\begin{figure}[t]
    \centering
    \begin{subfigure}[t]{0.48\linewidth}
        \centering
        \includegraphics[width=\linewidth]{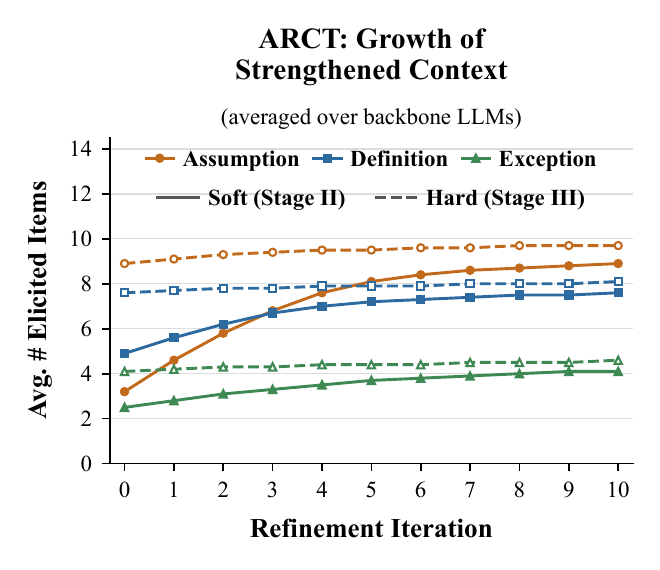}
        \label{fig:growth_arct}
    \end{subfigure}
    \hfill
    \begin{subfigure}[t]{0.48\linewidth}
        \centering
        \includegraphics[width=\linewidth]{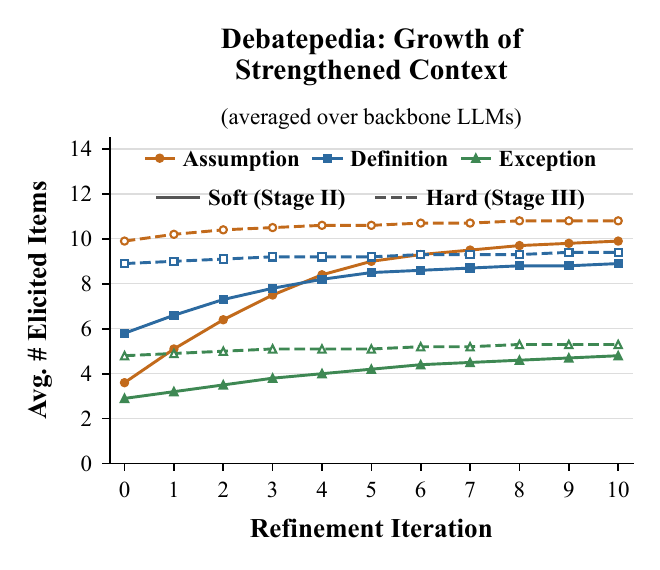}
        \label{fig:growth_debatepedia}
    \end{subfigure}
    \caption{Average number of assumptions, definitions, and exception conditions per instance across soft and hard refinement iterations on Debatepedia and ARCT.}
    \label{fig:growth}
\end{figure}
 
\textbf{Soft refinement constructs the strengthened context, while hard refinement repairs it.} Figure~\ref{fig:growth} tracks the average number of assumptions, definitions, and exception conditions per instance across refinement iterations. During soft refinement, assumptions and definitions grow steeply in the first five iterations, from an average of 3.6 to 9.0 and from an average of 5.8 to 8.5 on Debatepedia, which accounts for roughly 85\% of their total soft-phase growth: the framework is still proposing new items. After the fifth iteration both curves flatten while refinement continues, indicating that later soft rounds revise existing items rather than add new ones. Exceptions grow more slowly and almost linearly throughout (2.9 to 4.8 and 2.5 to 4.1), consistent with rebuttal conditions being added on demand, as counterexamples to a specific guard surface, rather than elicited up front. The hard phase barely changes the size of the context: ten hard iterations add at most 0.9 assumptions and 0.5 definitions or exceptions per instance on either dataset. Solver feedback thus drives local logical corrections to guards rather than proposing new elicited contexts.
 
\begin{table}[t]
\centering
\scriptsize
\setlength{\tabcolsep}{5pt}
\renewcommand{\arraystretch}{1.0}
\begin{tabular}{l cc}
\toprule
& VFR $\uparrow$ & Leak $\downarrow$ \\
\midrule
\multicolumn{3}{l}{\textsc{Debatepedia}} \\
\rowcolor{gray!10}
\method{} (full)                     & \textbf{76.52} & \textbf{2.07} \\
\quad w/o soft critique              & 36.03\,\textcolor{gray}{($-$40.5)} & 24.58 \\
\quad w/o assumptions ($A$)          & 53.04\,\textcolor{gray}{($-$23.5)} & 10.27 \\
\quad w/o definitions ($\Delta$)     & 62.35\,\textcolor{gray}{($-$14.2)} & 6.10 \\
\quad w/o exceptions ($X$)           & 73.68\,\textcolor{gray}{($-$2.8)}  & 2.67 \\
\addlinespace[4pt]
\multicolumn{3}{l}{\textsc{ARCT}} \\
\rowcolor{gray!10}
\method{} (full)                     & \textbf{81.00} & \textbf{1.82} \\
\quad w/o soft critique              & 41.50\,\textcolor{gray}{($-$39.5)} & 21.70 \\
\quad w/o assumptions ($A$)          & 49.50\,\textcolor{gray}{($-$31.5)} & 18.18 \\
\quad w/o definitions ($\Delta$)     & 71.50\,\textcolor{gray}{($-$9.5)}  & 7.14 \\
\quad w/o exceptions ($X$)           & 78.50\,\textcolor{gray}{($-$2.5)}  & 2.48 \\
\bottomrule
\end{tabular}
\caption{Ablation study with GPT-5.1 on Debatepedia and ARCT. The gray values in parentheses give the VFR drop relative to the full system. Results for the remaining backbones are reported in the Appendix \ref{sec:full-ablations}.}
\label{tab:ablation}
\end{table}
 
\textbf{The soft symbolic critique and the assumption layer carry most of the performance.} Table~\ref{tab:ablation} removes one component at a time with GPT-5.1. Removing the soft critique is the most damaging intervention and degrades validity and faithfulness at once: VFR falls by 40.5 points on Debatepedia and 39.5 on ARCT, leakage rises from 2.07\% to 24.58\% and from 1.82\% to 21.70\%. Among the elicitation layers, assumptions matter most: their removal costs 23.5 points on Debatepedia and 31.5 on ARCT, and the larger drop on ARCT reflects that its instances are constructed around a missing warrant, which is exactly what the assumption layer supplies. Definitions contribute a further 14.2 and 9.5 points, since term normalization is what lets guards and premise unify within a proof, while exceptions cost only 2.8 and 2.5 points in aggregate, as few accepted proofs need a rebuttal condition to close.
 
\begin{figure}[t]
    \centering
    \begin{subfigure}[t]{0.48\linewidth}
        \centering
        \includegraphics[width=\linewidth]{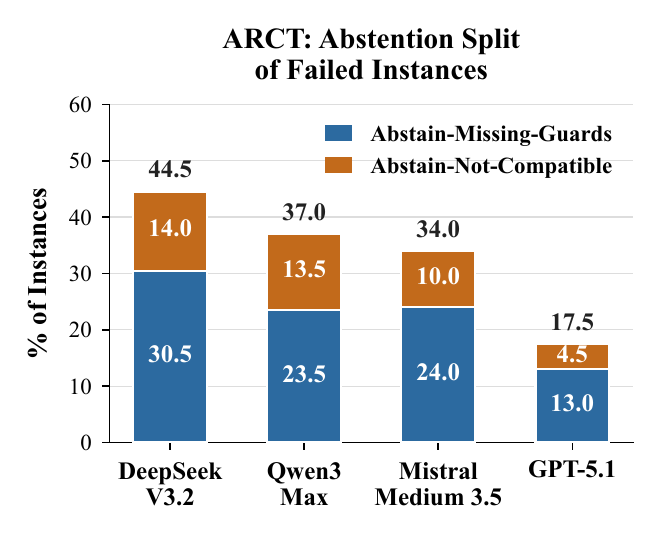}
        \label{fig:abstain_arct}
    \end{subfigure}
    \hfill
    \begin{subfigure}[t]{0.48\linewidth}
        \centering
        \includegraphics[width=\linewidth]{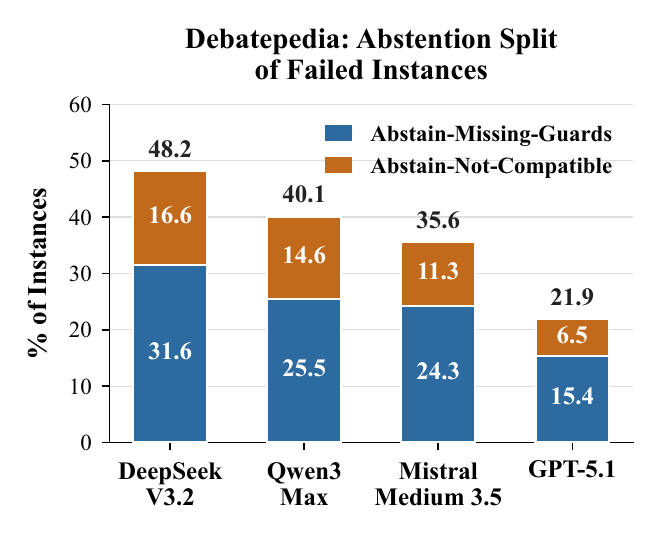}
        \label{fig:abstain_debatepedia}
    \end{subfigure}
    \caption{Decomposition of outcomes on abstain-missing-guards and abstain-not-compatible.}
    \label{fig:abstain}
\end{figure}
 
\textbf{Abstention is mostly a matter of missing guards, not of genuine incompatibility between premise and claim.} Figure~\ref{fig:abstain} decomposes the instances whose theory is not proved within the standard budget. We operationalize the two abstention states by granting each such instance ten additional hard iterations: cases that become provable are counted as abstain-missing-guards (AMG), and the remainder as abstain-not-compatible (ANC). Across backbones and datasets, 63.5--74.3\% of the unproven instances are AMG, so the dominant cause of abstention is an exhausted budget rather than an argument that resists explicitation; under the doubled budget, the solver-pass rate of GPT-5.1 would reach 93.5\% on Debatepedia and 95.5\% on ARCT. The remaining ANC core is small but persistent, ranging from 6.5\% of all instances for GPT-5.1 to 16.6\% for DeepSeek-V3.2 on Debatepedia and from 4.5\% to 14.0\% on ARCT, and it contracts monotonically as the backbone strengthens, indicating that part of what a weaker model reports as incompatible reflects the limits of its elicitation rather than of the argument.

\section{Related Work}

 
\noindent\textbf{Material inference, defeasibility, and implicit warrants.}
The conceptual background comes from inferentialist accounts of material inference \citep{brandomMakingItExplicit1994,brandomSayingDoingAnalytic2010,brandomArticulatingReasons2021,hlobil2024reasons} and from classical work on defeasible and non-monotonic reasoning \citep{reiter1980logic,mccarthy1980circumscription,pollock1987defeasible,dung1995acceptability}. In NLP, these concerns surface in argument reasoning benchmarks whose decisive step is the reconstruction of an implicit warrant \citep{habernal-etal-2018-argument,gupta-etal-2024-harnessing}. We borrow these distinctions to define what counts as a legitimate guard, but operationalize them computationally, as an autoformalization problem with explicit success and failure states.
 
\noindent\textbf{Argument mining, debate corpora, and support structure.}
Debatepedia and related corpora have been used to study argumentative support, contradiction, and structure over topic-bound debate text \citep{niculae2017argument,lawrenceArgumentMiningSurvey2020}, typically by predicting labels, retrieving support, or reconstructing warrants in natural language. Our setting is narrower but deeper: once a support relation is given, \method{} constructs the proof package that makes its hidden bridge explicit without altering the argumentative meaning of the original pair.
 
Closest to our work, solver-guided explanation verification for NLI \citep{quan-etal-2024-verification,quan-etal-2025-peirce} refines explanatory sentences that are already given while our method starts from debate premise-claim pairs, search for admissible guards, and report success only when the resulting proof is both solver-valid and contrastively faithful.
 
\section{Conclusion}
In this paper, we proposed \method{}, a neuro-symbolic framework that formulates the autoformalization of argumentative material inference as guard completion, turning implicit, non-monotonic support into explicit, deductively checkable inference. By combining context strengthening, symbolic soft critique, solver-guided hard refinement, and contrastive faithfulness checks, \method{} consistently outperforms state-of-the-art LLM-driven theorem proving on Debatepedia and ARCT across four backbone LLMs. In future work, we will extend guard completion to material inferences from a broader range of domains.

\bibliography{aaai2027}

\appendix
\setcounter{secnumdepth}{2}
\renewcommand{\thesection}{\Alph{section}}
\renewcommand{\thesubsection}{\thesection.\arabic{subsection}}
\renewcommand{\thesubsubsection}{\thesubsection.\arabic{subsubsection}}
\counterwithin{table}{section}
\counterwithin{figure}{section}
\counterwithin{equation}{section}
\counterwithin{lstlisting}{section}
\setcounter{table}{0}
\setcounter{figure}{0}
\setcounter{equation}{0}
\setcounter{lstlisting}{0}

\section{Detailed Experimental Configuration}
\label{sec:experimental-details}
 
\subsection{Datasets and Sampling}
\label{sec:dataset-details}
 
We evaluate on Debatepedia and the Argument Reasoning Comprehension Task (ARCT). The Debatepedia evaluation set contains 247 premise-claim pairs balanced across 19 topics, with 13 pairs per topic. The ARCT evaluation set contains 200 randomly sampled reason-claim pairs, which are treated as premise--claim pairs. The same fixed instances are used for every method and backbone so that all comparisons are paired at the instance level.

\begin{table*}[t]
\centering
\small
\begin{tabularx}{\textwidth}{@{}l c c >{\raggedright\arraybackslash}X@{}}
\toprule
Dataset & Instances & Topics & Construction used in this paper \\
\midrule
Debatepedia & 247 & 19 & Topic-balanced premise-claim pairs; 13 instances per topic \\
ARCT & 200 & -- & Fixed random sample of reason-claim pairs \\
\bottomrule
\end{tabularx}
\caption{Summary of the evaluation datasets.}
\label{tab:dataset-summary}
\end{table*}

\paragraph{Dataset selection.}
The evaluation targets argumentative support whose contextual basis remains partly implicit. A suitable dataset must provide a premise-claim relation for which the missing background commitments can be made explicit and assessed through formal verification. Debatepedia meets this requirement through argument pairs from online debates. Its extended resource distinguishes argumentative support from textual entailment \citep{cabrio2013natural}, making it relevant to the distinction between a materially supported argument and an inference whose support has been made formally explicit. This setting allows us to examine what contextual information is needed to turn the original support relation into a verifiable inference.

ARCT complements this setting by focusing directly on implicit warrants \citep{habernal-etal-2018-argument}. Its original task requires identifying the warrant that connects a reason to a claim. In our evaluation, each reason-claim pair serves as the input to guard completion. The system must construct the contextual support needed for formal verification, and success is assessed through the resulting proof and contrastive checks. The evaluation consequently concerns the explicit completion of the argument, without scoring the original warrant-selection task.

\paragraph{Topic coverage.}
The Debatepedia evaluation follows the 19-topic organisation of the extended resource \citep{cabrio2013natural}. These topics provide the basis for evaluating guard completion across different debate contexts. Assigning 13 premise-claim pairs to each topic gives every topic equal weight in the 247-instance evaluation set. Larger topics cannot dominate the aggregate results simply by contributing more instances. This balance concerns topic frequency and does not assume that all topics have the same reasoning difficulty.
 
\subsection{Backbone LLMs and Inference Settings}
\label{sec:llm-implementation}
 
We evaluate four backbone LLMs through their respective APIs. GPT-5.1 uses the identifier \texttt{gpt-5.1} \citep{openaiGPT512025}. Qwen3-Max uses \texttt{qwen3-max-preview} \citep{qwenTeamQwen3Max2025}. The identifiers for DeepSeek-V3.2 and Mistral Medium 3.5 are \texttt{deepseek-chat} \citep{deepseekV32025} and \texttt{mistral-medium-3-5} \citep{mistralMedium352026}, respectively. For GPT-5.1, we set reasoning\_effort to medium and do not specify a sampling temperature. We run Qwen3-Max and DeepSeek-V3.2 in non-thinking mode and set temperature=0.6 for Qwen3-Max, DeepSeek-V3.2, and Mistral Medium 3.5. We do not explicitly set
reasoning\_effort for Mistral Medium 3.5. For each backbone, all methods use identical decoding configurations for tasks shared across methods. Prompts are instantiated independently for each example, and outputs generated by one backbone are never used as demonstrations for any other backbone.
 
\begin{table*}[t]
\centering
\small
\begin{tabularx}{\textwidth}{@{}>{\raggedright\arraybackslash}p{0.32\textwidth} >{\centering\arraybackslash}p{0.15\textwidth} >{\raggedright\arraybackslash}X@{}}
\toprule
Stage or method & Maximum iterations & Termination condition \\
\midrule
Soft verification and refinement & 10 & PO1--PO4 pass, or budget exhausted \\
Hard verification and refinement & 10 & Isabelle/HOL proof succeeds, no admissible repair remains, or budget exhausted \\
Faithfulness refinement & 5 & Both contrastive tests pass, no admissible repair remains, or budget exhausted \\
PEIRCE & 10 & Theorem is proved, or budget exhausted \\
Extended abstention probe & +10 hard iterations & Proof succeeds or extended budget is exhausted \\
\bottomrule
\end{tabularx}
\caption{Refinement budgets and stopping conditions.}
\label{tab:iteration-budgets}
\end{table*}

\paragraph{Backbone selection.}
The backbone comparison evaluates whether the benefits of guard completion persist across different model families. The framework requires an LLM that can generate contextual support and express
natural language sentences in a formal representation. It must also respond to verification feedback during refinement. The four evaluated backbones support these roles through their respective APIs, allowing the same framework to be evaluated without depending on a single model family.

Using one model from each of four distinct families broadens the comparison beyond multiple versions of the same backbone. The principal comparison remains between methods using the same model, with the decoding settings for shared tasks held fixed. Repeating this comparison across backbones tests whether the contribution of guard completion remains consistent when the underlying LLM changes.

The selection supports a cross-family evaluation under the configurations reported below. It does not establish an exhaustive ranking of available LLMs. Other general-purpose backbones could be
evaluated through the same interface, while comparisons between additional versions of a single family would address sensitivity to model-version choice.
 
\subsection{Baseline Control Conditions}
\label{sec:baseline-details}
 
All methods share the same downstream autoformalisation and Isabelle/HOL verification pipeline. They differ only in how the missing inferential material is produced and whether it is iteratively revised. The Direct baseline receives no added contextual material. The chain-of-thought and Toulmin baselines generate their bridging context once. PEIRCE and \method{} receive theorem-prover feedback and may revise the added context within the common iteration budget. Every solver-passed candidate is subsequently evaluated with the same premise-contrast and claim-contrast tests used by \method{}.

 \paragraph{Baseline selection.}
The baselines are selected to examine how the treatment of missing context affects formal verification. Each comparison must produce an inference that can be assessed by the shared theorem prover. This requirement allows us to evaluate the contextual support provided by a method and to test whether a successful proof still depends on the original premise.

Direct autoformalisation provides the no-context control. It establishes what can be verified from the formalised premise and claim without constructing additional support. The comparison with
context-generating methods measures the contribution of making implicit commitments explicit within the same verification pipeline.

The 1-shot CoT baseline tests whether a short sequence of background sentences can supply the missing connection through general reasoning prompts \citep{wei2022chain}. Toulmin-style explication introduces an argumentation-specific account of that connection. Following the
motivation of \citet{gupta-etal-2024-harnessing}, it makes the warrant explicit and relates it to supporting background information. Its qualification and rebuttal condition also provide a way to express the circumstances under which the support applies. Our one-shot baseline adapts this idea to context generation for a fixed premise-claim pair. Comparing these two baselines examines whether
an explicit argumentation schema improves on a general sequence of background sentences when both contexts are generated in a single pass.

PEIRCE provides the iterative neuro-symbolic comparison \citep{quan-etal-2025-peirce}. In our setting, its explanatory sentences constitute added context that can be refined using theorem-prover feedback. Its inclusion allows us to compare \method{} with an approach that already benefits from iterative logical verification. The common limit of 10 verification and refinement iterations keeps the hard-refinement budget aligned between the two approaches.

\section{Autoformalisation Protocol}
\label{sec:autoformalisation}
 
\subsection{Representation Language}
\label{sec:representation-language}
 
Natural language sentences are translated into first-order logical forms with Neo-Davidsonian event variables \citep{parsons1990events}. Entity predicates introduce individuals, while verbal predicates introduce events. Thematic-role predicates connect events to their participants. Multiple verbal eventualities are represented by distinct event variables. Quantifiers must bind every variable in their scope. Polarity and connectives must match the source sentence. Predicates referring to the same entity, event, or relation must remain consistent throughout the instance.
 
The autoformaliser is realised as a three-step composition. For each sentence $s$, a syntactic-parsing step $\pi$ identifies the grammatical constituents. The parse records the subject with the main and auxiliary verbs. It also identifies objects and complements, together with adverbial modifiers. A formalisation step $F$ maps the parsed constituents into a Neo-Davidsonian logical form. A unification step $U_{\Sigma}$ then rewrites the forms over a shared predicate signature $\Sigma(S)$. Each concept has the same predicate symbol and arity throughout the instance.
\begin{equation}
\Phi(s) \;=\; U_{\Sigma(S)}\bigl(F(\pi(s))\bigr).
\label{eq:autoformalisation-pipeline}
\end{equation}
For an input set
\[
S = \{T,C\} \cup \Guard,
\]
the autoformaliser produces
\[
\Phi(S) = \{\Phi(s) \mid s \in S\},
\]
where the unification step is applied jointly over all elements of $S$ to enforce cross-sentence predicate consistency. Each selected guard becomes an axiom, while the formalised premise is introduced as a theorem assumption and the formalised claim as the proof goal.
 
As a running illustration, consider the Debatepedia instance whose premise is \emph{``Many solar energy systems are now price competitive with coal.''} The syntactic-parsing step returns the constituent structure
\begin{quote}\small
\texttt{Subject: Many solar energy systems}\\
\texttt{Linking Verb: are}\\
\texttt{Adverbial Modifier (Time): now}\\
\texttt{Subject Complement: price competitive with coal}
\end{quote}
from which the formalisation step produces
\begin{equation}
\begin{aligned}
\exists x\, y.\;&
\mathit{SolarEnergySystem}(x) \wedge \mathit{Many}(x)\\
&\wedge \mathit{Coal}(y) \wedge \mathit{Now}(x)\\
&\wedge \mathit{PriceCompetitiveWith}(x,y).
\end{aligned}
\label{eq:premise-lf}
\end{equation}
The associated claim \emph{``Solar energy is economically sound''} has a generic, non-episodic reading and is accordingly formalised with universal force,
\begin{equation}
\forall x.\; \mathit{SolarEnergy}(x) \longrightarrow \mathit{EconomicallySound}(x),
\label{eq:claim-lf}
\end{equation}
and an eventive guard such as \emph{``A solar energy system \ldots converts solar energy into usable electricity or heat''} introduces an explicit event variable with thematic roles, e.g.\ $\mathit{Convert}(e) \wedge \mathit{Agent}(e,y) \wedge \mathit{Patient}(e,z)$. The unification step guarantees, for instance, that $\mathit{SolarEnergy}$ and $\mathit{PriceCompetitiveWith}$ in Eq.~\eqref{eq:premise-lf} are the same symbols, with the same arities, that appear in the guards and in Eq.~\eqref{eq:claim-lf}.
 
\subsection{Formalisation of Exception-Management Items}
\label{sec:exception-compilation}
 
A defeasible rule family $d$ is formalised as a monotonic guarded rule using an abnormality predicate.
\begin{equation}
\forall k.\; \varphi_d(k) \wedge \neg \mathit{Ab}_d(k) \longrightarrow \psi_d(k).
\label{eq:guarded-rule}
\end{equation}
For a closed exception list, the abnormality condition is defined by
\begin{equation}
\mathit{Ab}_d(k) \longleftrightarrow
\chi_{d,1}(k) \vee \cdots \vee \chi_{d,n}(k).
\label{eq:abnormality-closed}
\end{equation}
For an open-ended list, only the direction from a listed exception to abnormality is asserted, avoiding the claim that the elicited exceptions are exhaustive. In the formalised theory, Eq.~\eqref{eq:autoformalisation-pipeline} is applied to the rule body $\varphi_d$ and head $\psi_d$. The abnormality disjuncts $\chi_{d,i}$ are autoformalised in the same way. These logical forms are then combined. A single guarded rule may contain several nested Davidsonian event descriptions, as illustrated by guard \texttt{G3} in Listing~\ref{lst:isabelle-example}.
 
\subsection{Automatic Quality Checks}
\label{sec:formalisation-checks}
 
Each logical form is checked against the following criteria before it is admitted to a theory.
\begin{enumerate}[leftmargin=2em]
  \item \textbf{Syntactic well-formedness.} The formula is complete and has balanced brackets. Its formalised representation uses legal Isabelle/HOL syntax.
  \item \textbf{Variable binding.} Every entity and event variable has a quantifier or local binder with the intended scope.
  \item \textbf{Predicate consistency.} A concept or event has a consistent predicate name and arity within the instance.
  \item \textbf{Semantic coverage.} The logical form represents the source sentence's content-bearing predicates and their arguments. It preserves modifiers and polarity, together with the logical connectives.
  \item \textbf{Role fidelity.} Semantic roles attach to the correct event and participant. The check covers agent and patient/theme roles, as well as recipient, location, and other roles in the sentence.
  \item \textbf{Inference direction.} Implications preserve the direction expressed by the source sentence. Causal links and defeasible rules follow the same requirement.
\end{enumerate}
 
\subsection{Faithfulness Evaluation of Autoformalisation}
\label{sec:autoformalisation-evaluation}
 
The main experiments share the autoformalisation component across all compared methods, so autoformalisation errors affect every system equally. We quantify the quality of this component using the two evaluation protocols of \citet{quan-etal-2025-faithful}. In \emph{informalisation}, an LLM translates each logical form back into natural language. The resulting sentence is compared with the source using the cosine similarity of their sentence embeddings,
\begin{equation}
\mathrm{faith}(s) \;=\;
\cos\bigl(\mathbf{e}(s),\, \mathbf{e}(\rho(\Phi(s)))\bigr),
\label{eq:informalisation}
\end{equation}
where $\rho$ denotes the informalisation map and $\mathbf{e}(\cdot)$ the sentence encoder.

In the \emph{human evaluation}, annotators inspect the generated Isabelle/HOL theories and assign each erroneous logical form to one of four categories. The \emph{syntax} category covers ill-formed or non-compiling formulas. An incorrect implication direction or strength is labelled \emph{implication}. The \emph{quantifier} category covers an incorrect quantifier or scope, while \emph{variable} covers unbound or incorrectly reused variables.
 
\begin{figure}[t]
\centering
\includegraphics[width=0.5\columnwidth]{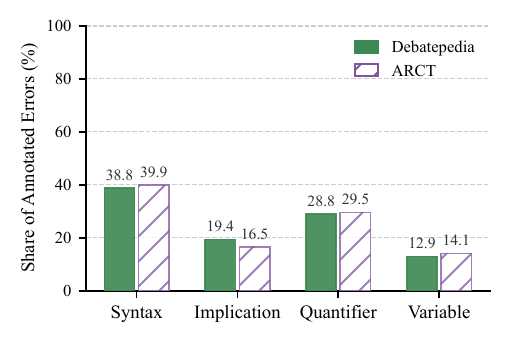}
\caption{Distribution of autoformalisation errors (syntax, implication, quantifier, variable) in the generated Isabelle/HOL theories, averaged over the four backbone LLMs. Annotation categories follow \citet{quan-etal-2025-faithful}.}
\label{fig:autoformalisation-errors}
\end{figure}

Table~\ref{tab:autoformalisation-faithfulness} reports the informalisation faithfulness over premise, claim, and guard sentences for each backbone; Figure~\ref{fig:autoformalisation-errors} shows the distribution of human-annotated error categories averaged over the four backbones on 300 randomly sampled logical forms. Faithfulness is high and stable across backbones (0.798--0.917), and the error mass concentrates on syntax errors, which are exactly the errors that the syntactic admission check of Section~\ref{sec:theory-construction} detects and repairs before verification; genuinely semantic errors (implication direction, quantification) account for a minority of cases.
 
\begin{table}[t]
\centering
\small
\setlength{\tabcolsep}{5pt}
\begin{tabular}{@{}l cc@{}}
\toprule
Backbone & Debatepedia & ARCT \\
\midrule
DeepSeek-V3.2       & 0.812 & 0.798 \\
Qwen3-Max           & 0.838 & 0.852 \\
Mistral Medium 3.5  & 0.861 & 0.869 \\
GPT-5.1             & 0.905 & 0.917 \\
\midrule
Avg.                & 0.854 & 0.859 \\
\bottomrule
\end{tabular}
\caption{Autoformalisation faithfulness (cosine similarity between each source sentence and the informalisation of its logical form, Eq.~\eqref{eq:informalisation}), computed over premise, claim, and guard sentences. The evaluation protocol follows \citet{quan-etal-2025-faithful}.}
\label{tab:autoformalisation-faithfulness}
\end{table}
 
\section{Isabelle/HOL Verification Details}
\label{sec:isabelle}
 
\subsection{Theory Construction}
\label{sec:theory-construction}
 
For each instance, the selected formal guards $G=\{g_1^\ast,\ldots,g_m^\ast\}$ are inserted as axioms. The theorem introduces the formal premise $T^\ast$ as an assumption and the formal claim $C^\ast$ as the goal. A proof succeeds only when Isabelle/HOL establishes
\begin{equation}
G \cup \{T^\ast\} \vdash C^\ast.
\label{eq:proof-obligation}
\end{equation}
The theory declares the base types \texttt{entity} and \texttt{event}. Each unified predicate receives a constant declaration, and each selected guard receives a named axiom \texttt{guard\_Gk}. The single theorem places $T^\ast$ in its \texttt{assumes} clause and $C^\ast$ in its \texttt{shows} clause. Listing~\ref{lst:isabelle-example} shows the (abbreviated) theory generated for the running example of Section~\ref{sec:representation-language}; the proof line is the tactic returned by Sledgehammer as described below.
 
\begin{lstlisting}[style=guard-isabelle-ascii,caption={Isabelle/HOL theory generated for the running example. Constant declarations, guard \texttt{G2}, and the inner event descriptions of guard \texttt{G3} (marked \texttt{[...]}) are abbreviated. The closing tactic \texttt{by blast} was found by Sledgehammer.},label={lst:isabelle-example}]
theory debatepedia_52_hard_0
  imports Main
begin
 
typedecl entity
typedecl event
 
consts
  SolarEnergySystem :: "entity => bool"
  PriceCompetitiveWith :: "entity => entity => bool"
  SolarEnergy :: "entity => bool"
  EnergySource :: "entity => bool"
  EconomicallySound :: "entity => bool"
  (* ... one constant per unified predicate ... *)
 
(* Guard G1: If many solar energy systems are now price
   competitive with coal, then every energy source that is
   solar energy is economically sound. *)
axiomatization where
  guard_G1: "(EX x y. SolarEnergySystem x & Many x & Coal y
              & Now x & PriceCompetitiveWith x y)
             --> (ALL z. (EnergySource z & SolarEnergy z)
              --> EconomicallySound z)"
 
(* Guard G3: compiled defeasible rule -- an energy source
   that generally delivers energy at prices competitive with
   coal for typical uses is economically sound, unless its
   apparent competitiveness is driven by major subsidies or
   hidden costs (the negated exception condition). Inner
   event descriptions are elided here as [...] . *)
axiomatization where
  guard_G3: "ALL x. (EnergySource x
             & (EX y u e. Energy y & Deliver e & Agent e x
                & Patient e y & Generally e & Price u
                & At e u & [...])
             & ~(EX z e1. ((Subsidy z & Major z)
                | (Cost z & Hidden z))
                & Drive e1 & Agent e1 z & [...]))
             --> EconomicallySound x"
 
(* Guard G4: Solar energy is radiant energy from sunlight,
   and everything that is solar energy is an energy source. *)
axiomatization where
  guard_G4: "ALL x. SolarEnergy x <-> (RadiantEnergy x
             & (EX y. Sunlight y & FromE x y) & EnergySource x)"
 
(* Guard G2 (definition of a solar energy system) omitted
   for space. *)
 
theorem hypothesis:
  (* Premise: Many solar energy systems are now price
     competitive with coal. *)
  assumes asm: "SolarEnergySystem x & Many x & Coal y
                & Now x & PriceCompetitiveWith x y"
  (* Claim: Solar energy is economically sound. *)
  shows "ALL x. SolarEnergy x --> EconomicallySound x"
  using assms guard_G1 guard_G4 by blast
 
end
\end{lstlisting}
 
Before any proof is attempted, the controller performs two automatic admission checks on the formalised theory. The \emph{syntax check} runs the theory through Isabelle and rejects it if the parser or type checker reports an error. The diagnostic and the affected lines are returned for LLM-based refinement, subject to a fixed number of refinement iterations. The \emph{consistency check} replaces the proof goal with $\mathit{False}$ and runs Sledgehammer. If
\begin{equation}
G \cup \{T^\ast\} \vdash \mathit{False}
\label{eq:consistency-check}
\end{equation}
is provable, the axiom set is contradictory. Any subsequent ``proof'' of $C^\ast$ would be vacuous, so the theory is rejected. The theorem prover output identifies the axioms used to prove the contradiction. Exactly those axioms are returned for LLM-based refinement. Only theories that pass both checks proceed to proof search.
 
\subsection{Proof Search with Sledgehammer}
\label{sec:proof-search}
 
Proof search proceeds in two modes. In the \emph{direct mode}, Sledgehammer \citep{paulson2010three} is invoked immediately on the stated goal of Eq.~\eqref{eq:proof-obligation}. The external automated provers search for a proof from the premise assumption and the guard axioms. On success, Sledgehammer reports a reconstructing tactic, such as \texttt{by blast} or \texttt{by (metis guard\_G1 guard\_G4)}. The controller splices the returned tactic into the theory in place of the Sledgehammer call and re-runs Isabelle to confirm that the reconstructed proof checks; the confirmed theory is what is stored and counted as solver-passed. The proof line of Listing~\ref{lst:isabelle-example} was obtained in this mode.
 
If no direct proof is found, the controller falls back to a \emph{proof-sketch mode}. An LLM-based proof-sketch function produces a structured Isar proof body that decomposes the entailment into a chain of intermediate facts
\begin{equation}
T^\ast \;=\; f_0 \;\Longrightarrow\; f_1 \;\Longrightarrow\; \cdots \;\Longrightarrow\; f_k \;=\; C^\ast,
\label{eq:proof-sketch}
\end{equation}
where each step names the load-bearing guard axioms it relies on and leaves its local justification as an \texttt{<ATP>} placeholder. The controller then discharges the placeholders from left to right. Each \texttt{<ATP>} is replaced by a Sledgehammer call, and the tactic Sledgehammer returns for that subgoal is substituted in place. The sketch succeeds only if every step is discharged and the completed proof is re-checked by Isabelle as a whole. If some step cannot be discharged, the failing step, its preceding comments, and the prover messages are recorded as the failure context for hard refinement. This two-mode design mirrors the observation that a single global Sledgehammer call suffices for short bridges, while multi-guard instances (e.g.\ those using a formalised defeasible rule together with a definition) benefit from an explicit intermediate decomposition.
 
\subsection{Failure Extraction and Hard Refinement}
\label{sec:failure-extraction}
 
When a theory fails, the controller records the failing theorem or proof step and the current goal state. It retains the relevant type or syntax diagnostics and Sledgehammer messages, together with the axioms supplied to that step. The LLM-based hard critique and refinement function $f_{\mathrm{refine}}^{\mathrm{hard}}$ receives this feedback alongside the current natural-language guards and their logical forms. The function classifies the failure under PO1--PO4 before proposing a refinement. It may correct an erroneous representation or add or refine a missing contextual item. Guard selection may also change, while the original premise and claim remain unchanged.

\section{Prompt Templates}
\label{sec:prompts}
 
Tables~\ref{tab:prompt_cot_baseline}--\ref{tab:prompt_faithfulness_repair}
list the prompt templates used in our experiments. The three baseline
prompts appear first, followed by the prompts used at each stage of the
pipeline. Curly-braced items such as \texttt{\{premise\}} are placeholders
instantiated at run time. The complete system and user prompts, including
all few-shot demonstrations, are provided in the submitted code.
 
\begin{guardprompt}{Baseline prompt: chain-of-thought background context.}{tab:prompt_cot_baseline}
\textbf{SYSTEM:} You are an expert in argumentative reasoning.
 
\smallskip
You will be given a context and a claim. The context is intended to support
the claim, but some of the connecting background information may be
implicit. Generate a short, ordered chain of background sentences that
makes the reasoning from the context to the claim explicit.
 
\smallskip
\textbf{Requirements:}\\
1. Treat the provided context as indispensable starting information.\\
2. The context and the generated background sentences, taken together, should support the claim.\\
3. The generated background sentences alone must not be sufficient to support the claim without the provided context.\\
4. Each background sentence should contribute one distinct general fact, rule, or intermediate connection.\\
5. Arrange the background sentences in the order in which they are needed to reach the claim.\\
6. Do not repeat or paraphrase the complete context.\\
7. Do not repeat, paraphrase, or directly state the complete claim.\\
8. Do not use a pair-specific shortcut such as ``If the context is true, then the claim is true.''\\
9. Do not introduce unrelated information, contradictory information, or a conclusion stronger than the provided claim.\\
10. Generate 2 to 4 concise declarative sentences, with one sentence on each line.\\
11. Return only the requested format without numbering, bullet points, or additional commentary.
 
\smallskip
\textbf{USER:} Here is one example:
 
\smallskip
\textit{Provided Context:} Television and texting are disturbing brains at a young age.\\
\textit{Provided Claim:} People are getting dumber.\\
\textit{Background Sentences:}\\
A disturbance to a person's brain at a young age impairs that person's later reasoning and learning abilities.\\
Impaired reasoning and learning abilities reduce a person's general cognitive ability over time.\\
A reduction in general cognitive ability over time constitutes a decline in intelligence.
 
\smallskip
Now generate the background sentences for the following instance:
 
\smallskip
\textit{Provided Context:} \texttt{\{premise\_sentence\}}\\
\textit{Provided Claim:} \texttt{\{hypothesis\_sentence\}}\\
\textit{Background Sentences:}
\end{guardprompt}
 
\begin{guardprompt}{Baseline prompt: explanatory-sentence generation.}{tab:prompt_explanation_baseline}
\textbf{SYSTEM:} You are an expert in natural language inference. A
hypothesis sentence and a premise sentence will be provided; however, the
premise sentence may not always be included. Please generate some
explanatory sentences to state how to infer the provided hypothesis
sentence.
 
\smallskip
\textbf{USER:} Here are some examples:
 
\smallskip
\textit{Provided Hypothesis Sentence:} Some children outside having fun and playing.\\
\textit{Provided Premise Sentence:} Three children in swimsuites are having fun outside near a white building.\\
\textit{Explanatory Sentences:}\\
Three children can be considered as some children, and they are children.\\
If some children are having fun outside, they are also playing outside.
 
\smallskip
\textit{Provided Hypothesis Sentence:} Azathioprine interferes with the assembly of proteins.\\
\textit{Explanatory Sentences:}\\
DNA contains genetic instructions for proteins, and RNA, synthesized from DNA, helps assemble the proteins.\\
The assembly of proteins is dependent on the synthesis of DNA and RNA.\\
Azathioprine interferes with DNA and RNA synthesis.
 
\smallskip
Do not generate explanatory sentences that just repeat or represent the
same meaning of the hypothesis sentence. The explanatory sentences should
form an inference chain to infer the hypothesis sentence and must not
repeat the hypothesis sentence.
 
\smallskip
Give me the explanatory sentences in the above example format:
 
\smallskip
\textit{Provided Hypothesis Sentence:} \texttt{\{hypothesis\_sentence\}}\\
\texttt{\{premise\_sentence\}}\\
\textit{Explanatory Sentences:}
\end{guardprompt}
 
\begin{guardprompt}{Baseline prompt: Toulmin-style context generation.}{tab:prompt_toulmin_baseline}
\textbf{SYSTEM:} You are an expert in argumentative reasoning.
 
\smallskip
You will be given a context and a claim. The context is intended to support
the claim, but the argumentative commitments connecting them may be
implicit. Generate a concise Toulmin-style context that makes the implicit
support relation explicit.
 
\smallskip
\textbf{The generated context should contain:}\\
1. One qualified bridging sentence that expresses the main general connection from the type of information stated in the context to the type of conclusion stated in the claim.\\
2. One backing sentence that provides a general fact or principle supporting the bridging connection.\\
3. When needed, one additional bridging sentence that completes an intermediate reasoning step or aligns the concepts used in the context and the claim.\\
4. One rebuttal condition describing a genuine circumstance under which the bridging connection may not apply.
 
\smallskip
\textbf{Requirements:}\\
1. The provided context and the generated Toulmin context sentences, taken together, should support the claim.\\
2. The generated sentences alone must not be sufficient to support the claim without the provided context.\\
3. The provided context must remain an indispensable starting point of the argument.\\
4. Express the appropriate inferential qualifier, such as ``generally'', ``normally'', ``typically'', or ``necessarily'', inside the main bridging sentence.\\
5. Use ``necessarily'' only when the connection is genuinely exceptionless; otherwise prefer a defeasible qualifier such as ``generally'' or ``normally''.\\
6. The bridging sentences must express general relations rather than facts that assert the claim about the current case.\\
7. The backing sentence must support the bridging relation and must not introduce new case-specific evidence that independently establishes the claim.\\
8. The rebuttal condition must describe only a possible exception; do not assert that the exception actually holds in the provided case.\\
9. Do not repeat or paraphrase the complete context.\\
10. Do not directly state or paraphrase the claim as a fact about the current case.\\
11. Do not use a pair-specific shortcut such as ``If the provided context is true, then the provided claim is true.''\\
12. Do not merely combine the complete context and claim into a conditional sentence.\\
13. Generate three to four concise sentences, with one sentence for each applicable argumentative role.\\
14. Return only the requested format without JSON, numbering, bullet points, or additional commentary.
 
\smallskip
\textbf{USER:} Here is one example:
 
\smallskip
\textit{Provided Context:} Television and texting are disturbing brains at a young age.\\
\textit{Provided Claim:} People are getting dumber.\\
\textit{Qualified Bridging Sentence:} Disturbing brain development at a young age generally impairs the later development of reasoning and learning abilities.\\
\textit{Backing Sentence:} Reasoning, learning, and memory depend on healthy brain development during childhood.\\
\textit{Additional Bridging Sentence:} People whose reasoning and learning abilities undergo a sustained decline become less intelligent over time.\\
\textit{Rebuttal Condition:} This connection may not apply when the disturbance is too minor to affect cognitive development or when its effects are fully offset by other influences.
 
\smallskip
Now generate the Toulmin context sentences for the following instance:
 
\smallskip
\textit{Provided Context:} \texttt{\{premise\_sentence\}}\\
\textit{Provided Claim:} \texttt{\{hypothesis\_sentence\}}\\
\textit{Toulmin Context Sentences:}
\end{guardprompt}
 
\begin{guardprompt}{Pipeline prompt: eliciting enthymematic assumptions.}{tab:prompt_assumptions}
\textbf{SYSTEM:} You are an expert in material inference and argument
reconstruction.
 
\smallskip
\textbf{USER:}\\
\textit{Premise:} \texttt{\{premise\}}\\
\textit{Claim:} \texttt{\{claim\}}
 
\smallskip
Elicit a concise set of implicit assumptions that are needed for the
premise to support the claim. Assumptions may include normality
conditions, domain restrictions, ceteris-paribus conditions, or an
implicit warrant.
 
\smallskip
\textbf{Constraints:}\\
1. Do not restate the claim or introduce a rule equivalent to the claim.\\
2. Do not add facts that contradict the premise.\\
3. Make each assumption independently understandable and formalizable.\\
4. State only assumptions that are potentially load-bearing.\\
5. Mark uncertainty explicitly rather than presenting a doubtful fact as universally true.
 
\smallskip
Return one item per line with a short justification of its role.
\end{guardprompt}
 
\begin{guardprompt}{Pipeline prompt: eliciting definitions and term-normalisation facts.}{tab:prompt_definitions}
\textbf{SYSTEM:} You are an expert in lexical semantics, definitions, and
formalization.
 
\smallskip
\textbf{USER:}\\
\textit{Premise:} \texttt{\{premise\}}\\
\textit{Claim:} \texttt{\{claim\}}\\
\textit{Candidate assumptions:} \texttt{\{assumptions\}}
 
\smallskip
Identify definitions or term-normalisation facts needed to align the
concepts used by the premise, claim, and assumptions.
 
\smallskip
\textbf{Constraints:}\\
1. Preserve the context-specific meaning of each term.\\
2. Do not broaden a term merely to make the claim provable.\\
3. Prefer a directional subsumption or equivalence only when it is semantically justified.\\
4. Do not introduce the claim, a paraphrase of the claim, or an answer-bearing definition.\\
5. Write each item as one formalizable sentence.
\end{guardprompt}
 
\begin{guardprompt}{Pipeline prompt: constructing defeasible rules and exceptions.}{tab:prompt_exceptions}
\textbf{SYSTEM:} You are an expert in defeasible reasoning and exception
management.
 
\smallskip
\textbf{USER:}\\
\textit{Premise:} \texttt{\{premise\}}\\
\textit{Claim:} \texttt{\{claim\}}\\
\textit{Assumptions:} \texttt{\{assumptions\}}\\
\textit{Definitions:} \texttt{\{definitions\}}
 
\smallskip
Construct only the defeasible rules needed to connect the premise to the
claim. For each rule, list concrete exception conditions that would block
its application.
 
\smallskip
\textbf{Constraints:}\\
1. Separate the rule antecedent, rule consequent, and exceptions.\\
2. Keep each rule local; do not use an unconditional rule whose consequent is the claim.\\
3. Exceptions must describe circumstances in which the rule is not reliable, not arbitrary reasons to reject the claim.\\
4. State whether the exception list is intended to be closed or open-ended.\\
5. Do not add an exception or rule that is not relevant to this inference.
\end{guardprompt}
 
\begin{guardprompt}{Pipeline prompt: translating sentences into logical forms.}{tab:prompt_autoformalisation}
\textbf{SYSTEM:} You are an expert in first-order logic and
Neo-Davidsonian event semantics.
 
\smallskip
\textbf{USER:}\\
\textit{Natural-language sentence:} \texttt{\{sentence\}}
 
\smallskip
Translate the sentence into one well-formed logical expression.
 
\smallskip
\textbf{Requirements:}\\
1. Introduce an event variable for every content-bearing verbal event.\\
2. Use distinct event variables for distinct events.\\
3. Bind every entity and event variable with the correct quantifier scope.\\
4. Preserve negation, modality, conditionals, conjunction, disjunction, and implication direction.\\
5. Use consistent predicates, arities, entities, and semantic roles.\\
6. Represent only content licensed by the sentence; do not add background knowledge or a predicate that is absent from its meaning.\\
7. Produce a form that can be compiled into Isabelle/HOL.
 
\smallskip
Return only the logical form and a compact predicate glossary.
\end{guardprompt}
 
\begin{guardprompt}{Pipeline prompt: selecting a minimal guard set.}{tab:prompt_selection}
\textbf{SYSTEM:} You are an expert in material inference, formal inference,
and argumentation theory.
 
\smallskip
\textbf{USER:}\\
\textit{Premise and formalization:} \texttt{\{premise\_and\_form\}}\\
\textit{Claim and formalization:} \texttt{\{claim\_and\_form\}}\\
\textit{Available assumptions, definitions, and exception-management items:} \texttt{\{strengthened\_context\}}
 
\smallskip
Select the smallest set of guards that jointly covers the missing
inferential material needed for the premise to support the claim.
 
\smallskip
\textbf{Constraints:}\\
1. Select only from the supplied context; do not invent a missing item.\\
2. Do not select the claim, a paraphrase of the claim, or an answer-bearing rule equivalent to the claim.\\
3. Every selected guard must be load-bearing: removing it should break the intended derivation.\\
4. The selected guards alone must not prove the claim without the premise.\\
5. Select zero or more items from each category according to necessity; do not pad the set merely to include every category.\\
6. If the context is insufficient, return FAIL and list repair candidates.
\end{guardprompt}
 
\begin{guardprompt}{Pipeline prompt: soft symbolic critique under PO1--PO4.}{tab:prompt_soft_verifier}
\textbf{SYSTEM:} You are a strict verifier of formalized material
inference.
 
\smallskip
\textbf{USER:}\\
\textit{Premise $T^{*}$:} \texttt{\{formal\_premise\}}\\
\textit{Claim $C^{*}$:} \texttt{\{formal\_claim\}}\\
\textit{Selected guards $G$:} \texttt{\{formal\_guards\}}
 
\smallskip
Evaluate the candidate under four proof obligations:\\
\textbf{PO1 Representation sanity:} syntax, types, arities, bindings, roles, and predicate normalization are coherent.\\
\textbf{PO2 Support adequacy:} the guards cover the missing inferential material and make relevant exceptions explicit.\\
\textbf{PO3 Strict support:} $G$ together with $T^{*}$ is sufficient to derive $C^{*}$.\\
\textbf{PO4 Dependence:} $G$ alone does not derive $C^{*}$, and every selected guard is necessary for the intended proof.
 
\smallskip
Do not silently repair the input while judging it. For every failed
obligation, identify the exact item responsible and provide either:
(a) a repair to an existing item; or (b) a description of a genuinely
missing item. Return PASS only when all four obligations pass.
\end{guardprompt}
 
\begin{guardprompt}{Pipeline prompt: refinement from Isabelle/HOL diagnostics.}{tab:prompt_hard_repair}
\textbf{SYSTEM:} You repair a formal theory using theorem-prover
diagnostics. Preserve the meaning of the original premise and claim.
 
\smallskip
\textbf{USER:}\\
\textit{Premise and claim:} \texttt{\{premise\_claim\}}\\
\textit{Current strengthened context and selected guards:} \texttt{\{current\_context\}}\\
\textit{Current logical forms:} \texttt{\{logical\_forms\}}\\
\textit{Isabelle/HOL diagnostic and failed proof step:} \texttt{\{prover\_feedback\}}
 
\smallskip
Diagnose the failure under PO1--PO4, then make the smallest admissible
repair. You may revise an erroneous logical form, repair or add a missing
context item, or change guard selection. You must not edit the original
premise or claim, assert the claim as a guard, or make the proof
independent of the premise. Return the complete updated context and guard
set.
\end{guardprompt}
 
\begin{guardprompt}{Pipeline prompt: premise- and claim-side contrastive probes.}{tab:prompt_contrastive}
\textbf{SYSTEM:} You generate hard contrastive tests for argumentative
support.
 
\smallskip
\textbf{USER:}\\
\textit{Original premise:} \texttt{\{premise\}}\\
\textit{Original claim:} \texttt{\{claim\}}
 
\smallskip
\textbf{Generate:}\\
\textbf{A1.} One same-topic counter-premise that should not support the original claim. It must be plausible and relevant to the broad topic, but it must not be a paraphrase or direct negation of the original premise.\\
\textbf{A2.} One nearby cross-subtopic claim that is not warranted by the original premise. It must be plausible and topically related, but it must test a materially different conclusion.
 
\smallskip
Explain briefly why each replacement should break the original support
relation. Do not generate an obviously nonsensical or contradictory probe.
\end{guardprompt}
 
\begin{guardprompt}{Pipeline prompt: repairing a solver-valid but leaky guard set.}{tab:prompt_faithfulness_repair}
\textbf{SYSTEM:} You repair a formally valid argument that failed a
contrastive faithfulness check.
 
\smallskip
\textbf{USER:}\\
\textit{Original premise, claim, guards, and proof:} \texttt{\{verified\_candidate\}}\\
\textit{Leaking contrastive theory and leak report:} \texttt{\{leak\_report\}}
 
\smallskip
\textbf{Revise the guard set so that:}\\
1. the original premise and revised guards still prove the original claim;\\
2. the guards alone do not prove the claim;\\
3. the counter-premise does not prove the claim with the guards;\\
4. the original premise does not prove the alternative claim with the guards; and\\
5. every guard remains semantically justified and load-bearing.
 
\smallskip
Narrow or condition over-broad rules rather than adding unrelated
blockers. Return the complete revised guard set, not only the changed
item.
\end{guardprompt}

\section{Complete Ablation Results}
\label{sec:full-ablations}
 
The main paper reports the GPT-5.1 ablation only.
Table~\ref{tab:ablation-full} reports the same ablation for every backbone
on both datasets. Removing the soft critique is the single most damaging
intervention for all four backbones, and the ordering
$A \succ \Delta \succ X$ in contribution size is stable across backbones
and datasets.
 
\begin{table*}[t]
\centering
\small
\setlength{\tabcolsep}{4.5pt}
\renewcommand{\arraystretch}{1.08}
\begin{tabular}{@{}l r@{\;}l r @{\hspace{1.8em}} r@{\;}l r@{}}
\toprule
 & \multicolumn{3}{c}{Debatepedia} & \multicolumn{3}{c}{ARCT} \\
\cmidrule(lr){2-4}\cmidrule(lr){5-7}
Variant & \multicolumn{2}{c}{VFR $\uparrow$} & Leak $\downarrow$
        & \multicolumn{2}{c}{VFR $\uparrow$} & Leak $\downarrow$ \\
\midrule
\multicolumn{7}{@{}l}{\textit{DeepSeek-V3.2}}\\
\quad \method{} (full)
  & \textbf{48.58} & & \textbf{\appzero6.25} & \textbf{53.50} & & \textbf{\appzero3.60}\\
\quad\quad w/o soft critique
  & 25.91 & \appdrop{22.7} & 29.67 & 29.00 & \appdrop{24.5} & 30.12\\
\quad\quad w/o assumptions ($A$)
  & 34.01 & \appdrop{14.6} & 16.83 & 36.50 & \appdrop{17.0} & 17.98\\
\quad\quad w/o definitions ($\Delta$)
  & 40.08 & \appdrop{8.5}  & 10.81 & 45.50 & \appdrop{8.0}  & \appzero9.90\\
\quad\quad w/o exceptions ($X$)
  & 45.34 & \appdrop{3.2}  & \appzero7.44 & 51.50 & \appdrop{2.0} & \appzero4.63\\
\addlinespace[0.45em]
\multicolumn{7}{@{}l}{\textit{Qwen3-Max}}\\
\quad \method{} (full)
  & \textbf{56.68} & & \textbf{\appzero5.41} & \textbf{61.00} & & \textbf{\appzero3.17}\\
\quad\quad w/o soft critique
  & 30.36 & \appdrop{26.3} & 29.25 & 33.50 & \appdrop{27.5} & 28.72\\
\quad\quad w/o assumptions ($A$)
  & 39.68 & \appdrop{17.0} & 16.24 & 42.00 & \appdrop{19.0} & 18.45\\
\quad\quad w/o definitions ($\Delta$)
  & 47.37 & \appdrop{9.3}  & \appzero8.59 & 52.00 & \appdrop{9.0} & \appzero8.77\\
\quad\quad w/o exceptions ($X$)
  & 54.25 & \appdrop{2.4}  & \appzero5.63 & 57.50 & \appdrop{3.5} & \appzero4.96\\
\addlinespace[0.45em]
\multicolumn{7}{@{}l}{\textit{Mistral Medium 3.5}}\\
\quad \method{} (full)
  & \textbf{62.75} & & \textbf{\appzero2.52} & \textbf{64.50} & & \textbf{\appzero2.27}\\
\quad\quad w/o soft critique
  & 37.65 & \appdrop{25.1} & 21.85 & 40.50 & \appdrop{24.0} & 24.30\\
\quad\quad w/o assumptions ($A$)
  & 46.56 & \appdrop{16.2} & 13.53 & 51.00 & \appdrop{13.5} & 12.82\\
\quad\quad w/o definitions ($\Delta$)
  & 56.28 & \appdrop{6.5}  & \appzero6.71 & 59.00 & \appdrop{5.5} & \appzero6.35\\
\quad\quad w/o exceptions ($X$)
  & 61.13 & \appdrop{1.6}  & \appzero3.82 & 63.00 & \appdrop{1.5} & \appzero3.08\\
\addlinespace[0.45em]
\multicolumn{7}{@{}l}{\textit{GPT-5.1}}\\
\quad \method{} (full)
  & \textbf{76.52} & & \textbf{\appzero2.07} & \textbf{81.00} & & \textbf{\appzero1.82}\\
\quad\quad w/o soft critique
  & 36.03 & \appdrop{40.5} & 24.58 & 41.50 & \appdrop{39.5} & 21.70\\
\quad\quad w/o assumptions ($A$)
  & 53.04 & \appdrop{23.5} & 10.27 & 49.50 & \appdrop{31.5} & 18.18\\
\quad\quad w/o definitions ($\Delta$)
  & 62.35 & \appdrop{14.2} & \appzero6.10 & 71.50 & \appdrop{9.5} & \appzero7.14\\
\quad\quad w/o exceptions ($X$)
  & 73.68 & \appdrop{2.8}  & \appzero2.67 & 78.50 & \appdrop{2.5} & \appzero2.48\\
\bottomrule
\end{tabular}
\caption{Complete ablation results for all backbone LLMs. VFR is the \vf{}
rate and Leak is the \rl{} rate, both in percent. Grey values give the
absolute VFR drop relative to the full \method{} pipeline. The GPT-5.1 rows
reproduce the ablation reported in the main paper.}
\label{tab:ablation-full}
\end{table*}
 
\section{End-to-End Example}
\label{sec:worked-examples}
 
This section traces Debatepedia instance \#197 through the full pipeline.
The example begins with context harvesting and follows the selected guards
through autoformalisation. It records soft and hard verification with their
refinements and concludes with the contrastive checks.
 
\subsection{Debatepedia \#197: Video-Game Play and Cognitive Benefits}
\label{sec:example-197}
 
\paragraph{Input.}
\begin{quote}
\textbf{Premise:} Moderate video-game play may have positive effects on
developing minds.\\
\textbf{Claim:} Video games can have cognitive benefits for young people.
\end{quote}
 
\paragraph{Strengthened context.}
Stage I harvests the following context inventory from the premise--claim
pair. Items later selected as guards are marked with their guard id.
 
\begin{description}[leftmargin=0pt,labelindent=0pt,labelwidth=0pt,labelsep=0pt,style=nextline,itemsep=5pt]
\item[Assumptions $A$]
\hfill\\
\textbf{A1} (bridging, implicit): Developing minds are the minds of young
people.\\
\textbf{A2} (semantic disambiguation, received standard; $\rightarrow$ G1):
Positive effects on developing minds are cognitive benefits for those
minds. \emph{Rationale: links the general phrase ``positive effects'' in
the premise to the specifically cognitive benefits asserted in the claim.}\\
\textbf{A3} (pair-specific shortcut, later discarded by selection): If
moderate video game play may have cognitive benefits for young people then
it is correct to say that video games can have cognitive benefits for
young people.
 
\item[Definitions $\Delta$]
\hfill\\
\textbf{D1} (moderate video-game play): Moderate video-game play is a
level of engagement with video games in terms of frequency and duration
that is neither minimal nor excessive and does not typically interfere
with healthy functioning.\\
\textbf{D2} (developing minds): Developing minds are the cognitive systems
of children and adolescents whose brain structure and mental functions are
still undergoing maturation.\\
\textbf{D3} (young people; $\rightarrow$ G2): Young people are children
and adolescents who have not yet reached full adulthood.\\
\textbf{D4} (cognitive benefits): Cognitive benefits are advantageous
changes such as improvements in mental processes including attention,
memory, reasoning, or problem solving.\\
\textbf{D5} (video-game play): Video-game play is the activity of
interacting with electronic games that display visual output on a screen
and respond to user input.\\
\textbf{D6} (positive effects): Positive effects are changes in outcomes
that are considered beneficial according to accepted psychological or
developmental indicators.
 
\item[Exception items $X$]
\hfill\\
\textbf{R1}: Moderate video-game play among young people typically
involves active and repeated cognitive engagement.
$\chi_{R1,1}$: the play consists mostly of passive watching or repetitive
actions that demand little mental effort.
$\chi_{R1,2}$: the level of use is wrongly classified as moderate when it
is actually minimal or excessive.\\
\textbf{R2}: Sustained cognitively engaging activities in young people
generally have the potential to produce positive cognitive effects.
$\chi_{R2,1}$: strong negative influences such as extreme stress, sleep
deprivation, or neurological damage prevent cognitively engaging
activities from producing positive cognitive effects.\\
\textbf{R3} ($\rightarrow$ G3, combined with $\chi_{R3,1}$): When an
activity for young people has a real possibility of producing positive
cognitive effects, it is usually acceptable to say that the activity can
have cognitive benefits for young people.
$\chi_{R3,1}$: the positive cognitive effects are so rare, small, or
fragile that ordinary speakers would not describe the activity as one that
can have cognitive benefits.
\end{description}
 
\paragraph{Selected guards.}
The one-time selector chooses a minimal initial guard set of three items
from the full inventory:
 
\begin{description}[leftmargin=0pt,labelindent=0pt,labelwidth=0pt,labelsep=0pt,style=nextline,itemsep=5pt]
\item[G1 (from A2)] Positive effects on developing minds are cognitive
benefits for those minds. \emph{Load-bearing because it converts the
premise's positive effects on developing minds into the claim-side
description of those same effects as cognitive benefits.}
\item[G2 (from D3)] Young people are children and adolescents who have not
yet reached full adulthood. \emph{Load-bearing because it aligns the
premise's target group (bearers of developing minds) with the claim's
young people.}
\item[G3 (from R3 $+ \chi_{R3,1}$)] The defeasible rule R3 combined with
its exception condition $\chi_{R3,1}$. \emph{Load-bearing because it
licenses the move from an activity's potential to produce positive
cognitive effects to the statement that the activity can have cognitive
benefits, while guarding against effects that are too rare, small, or
fragile.}
\end{description}
 
During the soft verification--refinement loop (below), a strict-support
failure triggers one guard refinement that adds four assumptions to the
workset, yielding guards G4--G7:
 
\begin{description}[leftmargin=0pt,labelindent=0pt,labelwidth=0pt,labelsep=0pt,style=nextline,itemsep=5pt]
\item[G4 (added, A3$'$)] Developing minds are the minds of young people.
\emph{Connects the mind affected in the premise to the young person it
belongs to.}
\item[G5 (added, A4)] Moderate video-game play with positive effects on
developing minds is an activity for young people that has a real
possibility of producing those effects. \emph{Lifts the premise facts into
the antecedent of the default rule G3.}
\item[G6 (added, A5)] The positive cognitive effects of moderate
video-game play are not rare, small, or fragile. \emph{Rules out the
exception condition $\chi_{R3,1}$ of G3, so the default applies.}
\item[G7 (added, A6)] If playing video games has a cognitive benefit for
young people, then video games have that cognitive benefit for young
people. \emph{Carries the conclusion from the playing activity to video
games, as the claim states.}
\end{description}
 
Hard verification later adds one further guard from prover feedback:
 
\begin{description}[leftmargin=0pt,labelindent=0pt,labelwidth=0pt,labelsep=0pt,style=nextline,itemsep=5pt]
\item[G8 (added by hard repair)] A cognitive benefit for a young person's
mind is a cognitive benefit for that young person. \emph{The failed proof
step needed to transfer the benefit from the developing mind to its
owner; no earlier guard licensed that transfer.}
\end{description}

\begin{table*}[t]
\centering
\small
\setlength{\tabcolsep}{5pt}
\begin{tabularx}{\textwidth}{@{}l >{\raggedright\arraybackslash}X >{\raggedright\arraybackslash}X@{}}
\toprule
Iteration & Diagnostic & Repair \\
\midrule
Soft 0 &
PO2 support adequacy passes on the selected guards G1--G3; PO3 strict
support fails: the premise yields a positive effect on a developing mind
and G1 recasts it as a cognitive benefit for that mind, but nothing
connects the developing mind to a young person, nothing lifts moderate
video-game play into the activity-for-young-people antecedent of the
default G3, nothing rules out G3's rare/small/fragile exception, and
nothing carries the conclusion from playing to video games. &
Guard refinement adds four assumptions in natural language (guards
G4--G7): the young-person bridge, the activity lifting, the exception
closure, and the play-to-game transfer. All Stage-II logical forms of the
workset are cleared and the selected package (premise, claim, and guards)
is re-autoformalised. \\
\midrule
Soft 1 &
All four proof obligations pass (PO1 representation sanity, PO2 support
adequacy, PO3 strict support, PO4 dependence); the loop stops with
SoftPass after one refinement. &
--- \\
\midrule
Hard 0 &
hard\_fail (proof-sketch mode, 179.3\,s): sledgehammer finds no proof at
the sketch step that strengthens the playing-benefit fact with the young
person, i.e.\ the transfer of the cognitive benefit from the developing
mind to the young person whose mind it is; no guard licensed that
transfer. &
Hard refinement keeps guards G1--G7 unchanged and adds guard G8: ``A
cognitive benefit for a young person's mind is a cognitive benefit for
that young person,''
$\forall z\,y\,u.\ (\mathit{Benefit}\,z \wedge \mathit{Cognitive}\,z
\wedge \mathit{For}\,z\,y \wedge \mathit{Mind}\,y \wedge
\mathit{Of}\,y\,u \wedge \mathit{YoungPerson}\,u) \longrightarrow
\mathit{For}\,z\,u$. \\
\midrule
Hard 1 &
hard\_pass (direct sledgehammer, 44.4\,s): cvc4, vampire, and
zipperposition all find proofs; the reconstructed tactic is
\texttt{using assms guard\_G1 guard\_G4 guard\_G7 guard\_G8 by blast}. &
--- \\
\bottomrule
\end{tabularx}
\caption{End-to-end refinement trace for Debatepedia \#197.}
\label{tab:example-trace}
\end{table*}

\paragraph{Formalisation and theory.}
The autoformalised premise $T^{*}$ and claim $C^{*}$ are:
 
\begin{lstlisting}[style=guard-logic,caption={Logical forms of the premise and claim.}]
T* (P1): ∃x y z e. VideoGamePlay(x) ∧ Moderate(x) ∧ Effect(z) ∧
         Positive(z) ∧ Mind(y) ∧ Developing(y) ∧ Have(e) ∧
         Agent(e, x) ∧ Patient(e, z) ∧ On(z, y)
 
C* (C1): ∃x y z e. VideoGame(x) ∧ Benefit(z) ∧ Cognitive(z) ∧
         YoungPerson(y) ∧ Have(e) ∧ Agent(e, x) ∧ Patient(e, z) ∧
         For(z, y)
\end{lstlisting}
 
The final hard-pass Isabelle/HOL theory, including all guard axioms
G1--G8 and the machine-found proof, is:
 
\Needspace{12\baselineskip}
\begin{lstlisting}[style=guard-isabelle,caption={Final hard-pass theory for Debatepedia \#197.}]
theory debatepedia_197_hard_1
imports Main
 
begin
 
typedecl entity
typedecl event
 
consts
  Effect :: "entity ⇒ bool"
  Positive :: "entity ⇒ bool"
  Mind :: "entity ⇒ bool"
  Developing :: "entity ⇒ bool"
  On :: "entity ⇒ entity ⇒ bool"
  Benefit :: "entity ⇒ bool"
  Cognitive :: "entity ⇒ bool"
  For :: "entity ⇒ entity ⇒ bool"
  YoungPerson :: "entity ⇒ bool"
  Child :: "entity ⇒ bool"
  Adolescent :: "entity ⇒ bool"
  Adulthood :: "entity ⇒ bool"
  Full :: "entity ⇒ bool"
  Reach :: "event ⇒ bool"
  Agent :: "event ⇒ entity ⇒ bool"
  Patient :: "event ⇒ entity ⇒ bool"
  Activity :: "entity ⇒ bool"
  PotentialToProduce :: "entity ⇒ entity ⇒ bool"
  Rare :: "entity ⇒ bool"
  Small :: "entity ⇒ bool"
  Fragile :: "entity ⇒ bool"
  Have :: "event ⇒ bool"
  Of :: "entity ⇒ entity ⇒ bool"
  VideoGamePlay :: "entity ⇒ bool"
  Moderate :: "entity ⇒ bool"
  VideoGame :: "entity ⇒ bool"
 
(* Guard G1: Positive effects on developing minds are cognitive benefits
   for those minds. *)
axiomatization where
  guard_G1: "∀z y. (Effect z ∧ Positive z ∧ Mind y ∧ Developing y ∧
    On z y) ⟶ (Benefit z ∧ Cognitive z ∧ For z y)"
 
(* Guard G2: Young people are children and adolescents who have not yet
   reached full adulthood. *)
axiomatization where
  guard_G2: "∀u. YoungPerson u ⟷ ((Child u ∨ Adolescent u) ∧
    ¬(∃v e. Adulthood v ∧ Full v ∧ Reach e ∧ Agent e u ∧ Patient e v))"
 
(* Guard G3: Defeasible rule: When an activity for young people has a real
   possibility of producing positive cognitive effects, it is usually
   acceptable to say that the activity can have cognitive benefits for
   young people. *)
axiomatization where
  guard_G3: "∀x. (Activity x ∧ (∃u. YoungPerson u ∧ For x u) ∧
    (∃z. Effect z ∧ Positive z ∧ Cognitive z ∧ PotentialToProduce x z) ∧
    (¬(∃z. Effect z ∧ Positive z ∧ Cognitive z ∧
       (Rare z ∨ Small z ∨ Fragile z)))) ⟶
    (∃z1 e1 u1. Benefit z1 ∧ Cognitive z1 ∧ Have e1 ∧ Agent e1 x ∧
     Patient e1 z1 ∧ YoungPerson u1 ∧ For z1 u1)"
 
(* Guard G4: Developing minds are the minds of young people. *)
axiomatization where
  guard_G4: "∀y. (Mind y ∧ Developing y) ⟶ (∃u. YoungPerson u ∧ Of y u)"
 
(* Guard G5: Moderate video-game play with positive effects on developing
   minds is an activity for young people that has a real possibility of
   producing those effects. *)
axiomatization where
  guard_G5: "∀x y z e u. (VideoGamePlay x ∧ Moderate x ∧ Effect z ∧
    Positive z ∧ Mind y ∧ Developing y ∧ On z y ∧ Have e ∧ Agent e x ∧
    Patient e z ∧ YoungPerson u ∧ Of y u) ⟶
    (Activity x ∧ For x u ∧ PotentialToProduce x z)"
 
(* Guard G6: The positive cognitive effects of moderate video-game play
   are not rare, small, or fragile. *)
axiomatization where
  guard_G6: "∀z. (Effect z ∧ Positive z ∧ Cognitive z) ⟶
    (¬Rare z ∧ ¬Small z ∧ ¬Fragile z)"
 
(* Guard G7: If playing video games has a cognitive benefit for young
   people, then video games have that cognitive benefit for young
   people. *)
axiomatization where
  guard_G7: "∀x z u e. (VideoGamePlay x ∧ Benefit z ∧ Cognitive z ∧
    Have e ∧ Agent e x ∧ Patient e z ∧ YoungPerson u ∧ For z u) ⟶
    (∃x1 e1. VideoGame x1 ∧ Have e1 ∧ Agent e1 x1 ∧ Patient e1 z ∧
     For z u)"
 
(* Guard G8: A cognitive benefit for a young person's mind is a cognitive
   benefit for that young person. *)
axiomatization where
  guard_G8: "∀z y u. (Benefit z ∧ Cognitive z ∧ For z y ∧ Mind y ∧
    Of y u ∧ YoungPerson u) ⟶ For z u"
 
theorem hypothesis:
  (* Premise: Moderate video-game play may have positive effects on
     developing minds. *)
  assumes asm: "VideoGamePlay x ∧ Moderate x ∧ Effect z ∧ Positive z ∧
    Mind y ∧ Developing y ∧ Have e ∧ Agent e x ∧ Patient e z ∧ On z y"
  (* Hypothesis: Video games can have cognitive benefits for young
     people. *)
  shows "∃x y z e. VideoGame x ∧ Benefit z ∧ Cognitive z ∧
    YoungPerson y ∧ Have e ∧ Agent e x ∧ Patient e z ∧ For z y"
  using assms guard_G1 guard_G4 guard_G7 guard_G8 by blast
 
end
\end{lstlisting}
 
\paragraph{Refinement trace.}
Table~\ref{tab:example-trace} reports the complete verification and
refinement trace.

\paragraph{Contrastive checks.}
All three Stage IV probes are run against the hard-pass theory with the
same guard axioms G1--G8 and direct sledgehammer.
 
\begin{description}[leftmargin=0pt,labelindent=0pt,labelwidth=0pt,labelsep=0pt,style=nextline,itemsep=5pt]
\item[Premise check (empty premise)] The \texttt{assumes asm} line is
removed from the theorem and only the \texttt{shows} goal is kept, so the
claim is attempted from the guard axioms alone. Sledgehammer finds no
proof (\emph{not provable}, 15.3\,s): the guard set does not entail the
claim without the premise, so the premise is genuinely load-bearing.
 
\item[Premise contrast check (same-topic counter-premise)] The generated
counter-premise ``Excessive video-game play is linked to attention
problems and poorer academic performance in developing minds'' replaces
the original premise while the claim is kept fixed. Its formalisation is
\begin{lstlisting}[style=guard-logic]
∃x y w v. VideoGamePlay(x) ∧ Excessive(x) ∧ AttentionProblem(w) ∧
  Performance(v) ∧ Academic(v) ∧ Poor(v) ∧ Mind(y) ∧ Developing(y) ∧
  LinkedTo(x, w) ∧ LinkedTo(x, v) ∧ In(w, y) ∧ In(v, y)
\end{lstlisting}
Sledgehammer finds no proof (\emph{not provable}, 45.3\,s): the guard set
does not license the claim from a harm-focused premise on the same topic.
 
\item[Hypothesis contrast check (alternative claim)] The generated
neighbor claim ``Excessive video-game play can have cognitive benefits
for young people'' replaces the original claim while the premise is kept
fixed. Its formalisation is
\begin{lstlisting}[style=guard-logic]
∃x y z e. VideoGamePlay(x) ∧ Excessive(x) ∧ Benefit(z) ∧ Cognitive(z) ∧
  YoungPerson(y) ∧ Have(e) ∧ Agent(e, x) ∧ Patient(e, z) ∧ For(z, y)
\end{lstlisting}
Sledgehammer finds no proof (\emph{not provable}, 44.9\,s): the premise
about \emph{moderate} play does not warrant a conclusion about
\emph{excessive} play under the same guards.
\end{description}
 
Since the original theorem is provable and all three contrastive probes
are not provable, the instance is verified \vf{}.


\end{document}